\documentclass[sw]{iosart2x}

\usepackage{hyperref}
\usepackage{dcolumn}
\usepackage{algorithm}
\usepackage{algorithmic}
\usepackage{multirow,tabularx}
\usepackage{xcolor}
\usepackage{pdflscape}
\usepackage{afterpage}
\usepackage{capt-of}

\usepackage{lipsum}

\newcolumntype{d}[1]{D{.}{.}{#1}}

\firstpage{1}
\lastpage{5}
\volume{1}
\pubyear{0000}

\begin{document}
\begin{frontmatter} 

%
\title{Hate Speech Detection: A Solved Problem? The Challenging Case of Long Tail on Twitter}

\runningtitle{Hate Speech Detection - the Difficult Long-tail Case of Twitter}

\author{\fnms{Ziqi} \snm{Zhang}\thanks{Correspondence author}}\\
\runningauthor{Z. Zhang and L. Luo}
\address{Information School, University of Sheffield, Regent Court, 211 Portobello, Sheffield, S1 4DP, UK\\
	E-mail: ziqi.zhang@sheffield.ac.uk}

\author{\fnms{Lei} \snm{Luo}}\\
\address{College of Pharmaceutical Science, Southwest University, Chongqing, 400716, China\\
	E-mail: drluolei@swu.edu.cn}

\begin{review}{editor}
\reviewer{\fnms{Luis Espinosa Anke} \address{\orgname{Cardiff University}, \cny{UK}}}
\reviewer{\fnms{Thierry Declerck} \address{\orgname{DFKI GmbH}, \cny{Germany}}}
\reviewer{\fnms{Dagmar Gromann} \address{\orgname{Technical University Dresden}, \cny{Germany}}}
\end{review}
\begin{review}{solicited}
\reviewer{\fnms{Martin Riedl} \address{\orgname{Universitat Stuttgart}, \cny{Germany}}}
\reviewer{\fnms{Armand Vilalta} \address{\orgname{Barcelona Supercomputing Centre}, \cny{Spain}}}
\reviewer{\fnms{Anonymous} \address{\orgname{Unknown}, \cny{Unknown}}}
\end{review}
\begin{review}{open}
\reviewer{\fnms{Martin Riedl} \address{\orgname{Universitat Stuttgart}, \cny{Germany}}}
\reviewer{\fnms{Armand Vilalta} \address{\orgname{Barcelona Supercomputing Centre}, \cny{Spain}}}
\end{review}

\begin{abstract}
In recent years, the increasing propagation of hate speech on social media and the urgent need for effective counter-measures have drawn significant investment from governments, companies, and researchers. A large number of methods have been developed for automated hate speech detection online. This aims to classify textual content into non-hate or hate speech, in which case the method may also identify the targeting characteristics (i.e., types of hate, such as race, and religion) in the hate speech. However, we notice significant difference between the performance of the two (i.e., non-hate v.s. hate). In this work, we argue for a focus on the latter problem for practical reasons. We show that it is a much more challenging task, as our analysis of the language in the typical datasets shows that hate speech lacks unique, discriminative features and therefore is found in the `long tail' in a dataset that is difficult to discover. We then propose Deep Neural Network structures serving as feature extractors that are particularly effective for capturing the semantics of hate speech. Our methods are evaluated on the largest collection of hate speech datasets based on Twitter, and are shown to be able to outperform the best performing  method by up to 5 percentage points in macro-average F1, or 8 percentage points in the more challenging case of identifying hateful content.

\end{abstract}

\end{frontmatter}

\section{Introduction}\label{sec:intro}

The exponential growth of social media such as Twitter and community forums has revolutionised communication and content publishing, but is also increasingly exploited for the propagation of hate speech and the organisation of hate-based activities \cite{Badjatiya2017,Burnap2015}. The anonymity and mobility afforded by such media has made the breeding and spread of hate speech -- eventually leading to hate crime -- effortless in a virtual landscape beyond the realms of traditional law enforcement. 

The term `hate speech' was formally defined as `any communication that disparages a person or a group on the basis of some characteristics (to
be referred to as \textbf{types of hate} or \textbf{hate classes}) such as race, colour, ethnicity, gender, sexual orientation, nationality, religion, or other characteristics' \cite{Nockleby2000}. In the UK, there has been significant increase of hate speech towards the migrant and Muslim communities following recent events including leaving the EU, the Manchester and the London attacks \cite{TheGuardian2017}. 
In the EU, surveys and reports focusing on young people in the EEA (European Economic Area) region show rising hate speech and related crimes based on religious beliefs, ethnicity, sexual orientation or gender, as 80\% of respondents have encountered hate speech online and 40\% felt attacked or threatened \cite{EEA2017}. Statistics also show that in the US, hate speech and crime is on the rise since the Trump election \cite{Okeowo2016}. The urgency of this matter has been increasingly recognised, as a range of international initiatives have been launched towards the qualification of the problems and the development of counter-measures 
\cite{Galiardone2015}.

Building effective counter measures for online hate speech requires as the first step, identifying and tracking hate speech online.  
For years, social media companies such as Twitter, Facebook, and YouTube have been investing hundreds of millions of euros every year on this task  \cite{Lomas2015,TheGuardian2016,Gamback2017}, but are still being criticised for not doing enough. This is largely because such efforts are primarily based on manual moderation to identify and delete offensive materials. The process is labour intensive, time consuming, and not sustainable or scalable in reality \cite{Chen2012,Gamback2017,Waseem2016a}. 

A large number of research has been conducted in recent years to develop automatic  methods for hate speech detection in the social media domain. These typically employ semantic content analysis techniques built on Natural Language Processing (NLP) and Machine Learning (ML) methods, both of which are core pillars of the Semantic Web research. The task typically involves classifying textual content into non-hate or hateful, in which case
it may also identify the types of the hate speech. Although current methods have reported promising results, we notice that their evaluations are largely biased towards detecting content that is non-hate, as opposed to detecting and classifying real hateful content. A limited number of studies \cite{Burnap2015,Park2017} have shown that, for example, state of the art methods that detect sexism messages can only obtain an F1 of between 15 and 60 percentage points lower than detecting non-hate messages. These results suggest that it is much harder to detect hateful content and their types than non-hate\footnote{Even in a binary setting of the task (i.e., either a message is hate or not), the high accuracy obtainable on detecting non-hate does not automatically translate to high accuracy of the other task due to the highly imbalanced nature in such datasets, as we shall show later.}. However, from a practical point of view, we argue that the ability to correctly (Precision) and thoroughly (Recall) detect and identify specific types of hate speech is more desirable. For example, social media companies need to flag up hateful content for moderation, while law enforcement need to identify hateful messages and their nature as forensic evidence. 


Motivated by these observations, our work makes two major contributions to the research of online hate speech detection. \textbf{First}, we conduct a data analysis to quantify and qualify the linguistic characteristics of such content on the social media, in order to understand the challenging case of detecting hateful content compared to non-hate. By comparison, we show that hateful content exhibits a `long tail' pattern compared to non-hate due to their lack of unique, discriminative linguistic features, and this makes them very difficult to identify using conventional features widely adopted in many language-based tasks. \textbf{Second}, we propose Deep Neural Network (DNN) structures that are empirically shown to be very effective feature extractors for identifying specific types of hate speech. These include two DNN models inspired and adapted from literature of other Machine Learning tasks: one that simulates skip-gram like feature extraction based on modified Convolutional Neural Networks (CNN), and another that extracts orderly information between features using Gated Recurrent Unit networks (GRU).

Evaluated on the largest collection of English Twitter datasets, we show that our proposed methods can outperform state of the art methods by up to 5 percentage points in macro-average F1, or 8 percentage points in the more challenging task of detecting and classifying hateful content. Our thorough evaluation on all currently available public Twitter datasets sets a new benchmark for future research in this area. And our findings encourage future work to take a renewed perspective, i.e., to consider the challenging case of long tail.

The remainder of this paper is structured as follows. Section \ref{related_work} reviews related work on hate speech detection and other relevant fields; Section \ref{dataset_analysis} describes our data analysis to understand the challenges of hate speech detection on Twitter; Section \ref{methodology} introduces our methods; Section \ref{experiment} presents experiments and results; and finally Section \ref{conclusion} concludes this work and discusses future work.
\section{Related Work}\label{related_work}

\subsection{Terminology and Scope}
\begin{sloppypar}
Recent years have seen an increasing number of research on hate speech detection as well as other related areas. As a result, the term `hate speech' is often seen to co-exist or become mixed with other terms such as `offensive', `profane', and `abusive languages', and `cyberbullying'. To distinguish them, we identify that hate speech 1) targets individual or groups on the basis of their characteristics; 2) demonstrates a clear intention to incite harm, or to promote hatred; 3) may or may not use offensive or profane words. For example: \texttt{`Assimilate? No they all need to go back to their own countries. \#BanMuslims Sorry if someone disagrees too bad.'}
\end{sloppypar}

\begin{sloppypar}
In contrast, \texttt{`All you perverts (other than me) who posted today, needs to leave the O Board. Dfasdfdasfadfs'} is an example of abusive language, which often bears the purpose of insulting individuals or groups, and can include hate speech, derogatory and offensive language \cite{Nobata2016}. \texttt{`i spend my money how i want bitch its my business'} is an example of offensive or profane language, which is typically characterised by the use of swearing or curse words. \texttt{`Our class prom night just got ruined because u showed up. Who invited u anyway?'} is an example of bullying, which has the purpose to harass, threaten or intimidate typically individuals rather than groups.
\end{sloppypar}

In the following, we cover state of the art in all these areas with a focus on hate speech\footnote{We will indicate explicitly where works address a related problem rather than hate speech.}. Our methods and experiments will only address hate speech, due to both dataset availability and the goal of this work. 

\subsection{Methods of Hate Speech Detection and Related Problems}
Existing methods primarily cast the problem as a supervised document classification task \cite{Schmidt2017}. These can be divided into two categories: one relies on manual feature engineering that are then consumed by algorithms such as SVM, Naive Bayes, and Logistic Regression \cite{Burnap2015,Davidson2017,Djuric2015,Greevy2004,Kwok2013,Mehdad2016,Warner2012,Waseem2016a,Waseem2016b,Xiang2012,Yuan2016} (\textbf{classic methods}); the other represents the more recent deep learning paradigm that employs neural networks to automatically learn multi-layers of abstract features from raw data \cite{Gamback2017,Nobata2016,Park2017,Vigna2017} (\textbf{deep learning methods}). 
\newline
\newline
\noindent \textbf{Classic methods} require manually designing and encoding features of data instances into feature vectors, which are then directly used by classifiers. 

Schmidt et al. \cite{Schmidt2017} summarised several types of features used in the state of the art. \textit{Simple surface features} such as bag of words, word and character n-grams have been used as fundamental features in hate speech detection \cite{Burnap2015,Burnap2016,Davidson2017,Greevy2004,Kwok2013,Vigna2017,Warner2012,Waseem2016a,Waseem2016b}, as well as other related tasks such as the detection of offensive and abusive content \cite{Chen2012,Mehdad2016,Nobata2016}, discrimination \cite{Yuan2016}, and cyberbullying \cite{Zhong2016}. 
Other surface features can include URL mentions, hashtags, punctuations, word and document lengths, capitalisation, etc \cite{Chen2012,Davidson2017,Nobata2016}. \textit{Word generalisation} includes the use of low-dimensional, dense vectorial word representations usually learned by clustering  \cite{Warner2012}, topic modelling \cite{Xiang2012,Zhong2016} 
, and word embeddings \cite{Djuric2015,Nobata2016,Vigna2017,Yuan2016} from unlabelled corpora. Such word representations are then used to construct feature vectors of messages. \textit{Sentiment analysis} makes use of the degree of polarity expressed in a message \cite{Burnap2015,Davidson2017,Gitari2015,Vigna2017}. 
\textit{Lexical resources} are often used to look up specific negative words (such as slurs, insults, etc.) in messages 
\cite{Burnap2015,Gitari2015,Nobata2016,Xiang2012}. 
\textit{Linguistic features} utilise syntactic information such as Part of Speech (PoS) and certain dependency relations as features \cite{Burnap2015,Chen2012,Davidson2017,Gitari2015,Zhong2016}. 
\textit{Meta-information} refers to data about messages, such as gender identity of a user associated with a message \cite{Waseem2016a,Waseem2016b}, or high frequency of profane words in a user's post history \cite{Xiang2012,Dadvar2013}. In addition, \textit{Knowledge-Based features} such as messages mapped to stereotypical concepts in a knowledge base \cite{Dinakar2012} and \textit{multimodal information} such as image captions and pixel features \cite{Zhong2016} were used in cyberbullying detection but only in very confined context \cite{Schmidt2017}. 

In terms of classifiers, existing methods are predominantly supervised. Among these, Support Vector Machines (SVM) is the most popular algorithm \cite{Burnap2015,Chen2012,Davidson2017,Greevy2004,Mehdad2016,Warner2012,Xiang2012,Yuan2016}, while other algorithms such as Naive Bayes \cite{Chen2012,Davidson2017,Kwok2013,Mehdad2016,Yuan2016}, Logistic Regression \cite{Davidson2017,Djuric2015,Mehdad2016,Waseem2016a,Waseem2016b}, and Random Forest \cite{Davidson2017,Xiang2012} are also used. 
\newline
\newline
\noindent \textbf{Deep learning based methods} employ deep artificial neural networks to learn abstract feature representations from input data through its multiple stacked layers for the classification of hate speech. The input can be simply the raw text data, or take various forms of feature encoding, including any of those used in the classic methods. However, the key difference is that in such a model the input features may not be directly used for classification. Instead, the multi-layer structure can be used to learn from the input, new abstract feature representations that prove to be more effective for learning. For this reason, deep learning based methods typically shift its focus from manual feature engineering to the network structure, which is carefully designed to automatically extract useful features from a simple input feature representation. Indeed we notice a clear trend in the literature that shifts towards the adoption of deep learning based methods and studies have also shown them to perform better than classic methods on this task \cite{Gamback2017,Park2017}. Note that this categorisation excludes those methods \cite{Djuric2015,Mehdad2016,Yuan2016} that used DNN to learn word or text embeddings and subsequently apply another classifier (e.g., SVM, logistic regression) to use such embeddings as features for classification. Instead, we focus on DNN methods that perform the classification task itself. 

To the best of our knowledge, methods of this category include \cite{Badjatiya2017,Gamback2017,Park2017,Vigna2017,Zhang2018}, all of which used simple word and/or character based one-hot encoding as input features to their models, while Vigna et al. \cite{Vigna2017} also used word polarity. The most popular network architectures are Convolutional Neural Network (CNN) and Recurrent Neural Network (RNN), typically Long Short-Term Memory network (LSTM). In the literature, CNN is well known as an effective network to act as `feature extractors', whereas RNN is good for modelling orderly sequence learning problems \cite{Ordonez2016}. In the context of hate speech classification, intuitively, CNN extracts word or character combinations \cite{Badjatiya2017,Gamback2017,Park2017} (e.g., phrases, n-grams), RNN learns word or character dependencies (orderly information) in tweets \cite{Badjatiya2017,Vigna2017}. 

In our previous work \cite{Zhang2018}, we showed benefits of combining both structures in such tasks by using a hybrid CNN and GRU (Gated Recurrent Unit) structure. This work largely extends it in several ways. First, we adapt the model to multiple CNN layers; second, we propose a new DNN architecture based on the idea of extracting skip-gram like features for this task; third, we conduct data analysis to understand the challenges in hate speech detection due to the linguistic characteristics in the data; and finally, we perform an extended evaluation of our methods, particularly their capability on addressing these challenges.

\subsection{Evaluation of Hate Speech Detection Methods}
Evaluation of the performance of hate speech (and also other related content) detection typically adopts the classic Precision, Recall and F1 metrics. Precision measures the percentage of true positives among the set of hate speech messages identified by a system; Recall measures the percentage of true positives among the set of real hate speech messages we expect the system to capture (also called `\textbf{ground truth}' or `\textbf{gold standard}'), and F1 calculates the harmonic mean of the two. The three metrics are usually applied to each class in a dataset, and often an aggregated figure is computed either using \textbf{micro-average} or \textbf{macro-average}. The first sums up the individual true positives, false positives, and false negatives identified by a system regardless of different classes to calculate overall Precision, Recall and F1 scores. The second takes the average of the Precision, Recall and F1 on different classes. 

Existing studies on hate speech detection have primarily reported their results using micro-average Precision, Recall and F1 \cite{Badjatiya2017,Gamback2017,Park2017,Waseem2016a,Waseem2016b,Zhang2018}. The problem with this is that in an unbalanced dataset where instances of one class (to be called the `dominant class') significantly out-number others (to be called `minority classes'), micro-averaging can mask the real performance on minority classes. Thus a significantly lower or higher F1 score on a minority class (when compared to the majority class) is unlikely to cause significant change in micro-F1 on the entire dataset. As we will show in Section \ref{dataset_analysis}, hate speech detection is a typical task dealing with extremely unbalanced datasets, where real hateful content only accounts for a very small percentage of the entire dataset, while the large majority is non-hate but exhibits similar linguistic characteristics to hateful content. As argued before, practical applications often need to focus on detecting hateful content and identifying their types. In this case, reporting micro F1 on the entire dataset will not properly reflect a system's ability to deal with hateful content as opposed to non-hate. Unfortunately, only a very limited number of work has reported performance on a per-class basis \cite{Burnap2016,Park2017}. As an example, when compared to the micro F1 scores obtained on the entire dataset, the highest F1 score reported for detecting sexism messages is 47 percentage points lower in \cite{Burnap2016} while 11 points lower in \cite{Park2017}. This has largely motivated our study to understand what causes hate speech to be so difficult to classify from a linguistic point of view, and to evaluate hate speech detection methods by giving more focus on their capability of classifying real hateful content.




\section{Dataset Analysis - the Case of Long Tail}\label{dataset_analysis}
We first start with an analysis of typical datasets used in the studies of hate speech detection on Twitter. From this we show the very unbalanced nature of such data, and compare the linguistic characteristics of hate speech against non-hate to discuss the challenge of detecting and classifying hateful content. 

\subsection{Public Twitter Datasets} \label{dataset}
We use the collection of publicly available English Twitter datasets previously compiled in our work \cite{Zhang2018}. To our knowledge, this is the largest set (in terms of tweets) of Twitter based dataset used in hate speech detection. This includes seven datasets published in previous research. And all of these were collected based on the principle of keyword or hashtag filtering from the public Twitter stream. \textbf{DT} consolidates the dataset by \cite{Davidson2017} into two types, `hate' and `non-hate'. The tweets do not have a focused topic but were collected using a controlled vocabulary of abusive words. \textbf{RM} contains `hate' and `non-hate' tweets focused on refugee and muslim discussions. \textbf{WZ} is initially published by \cite{Waseem2016b} and contains `sexism', `racism', and `non-hate'; the same authors created another smaller dataset annotated by domain experts and amateurs separately. The authors showed that the two sets of annotations were different as the supervised classifiers obtained different results on them. We will use \textbf{WZ-S.amt} to denote the dataset annotated by amateurs, and \textbf{WZ-S.exp} to denote the dataset annotated by experts. \textbf{WZ-S.gb} merges the WZ-S.amt and WZ-S.exp datasets by taking the majority vote from both amateur and expert annotations where the expert was given double weights \cite{Gamback2017}. \textbf{WZ.pj} combines the WZ and the WZ-S.exp datasets \cite{Park2017}. All of the WZ-S.amt, WZ-S.exp, WZ-S.gb, and WZ.pj datasets contain `sexism', `racism', and `non-hate' tweets, but also added a `both' class that includes tweets considered to be both `sexism' and `racism'. However, there are only several handful of instances of this class and they were found to be insufficient for model learning. Therefore, following \cite{Park2017} we exclude this class from these datasets. Table \ref{tab_datasets} summarises the statistics of these datasets.

\begin{table}[thb]
    \centering
    \begin{tabular}{p{1.3cm}|c|p{3cm}}
        \hline
        \textbf{Dataset} 		& \textbf{\#Tweets} 	& \textbf{Classes} (\%)\\
				\hline
        WZ							  & 16,093						  & racism (12\%) sexism (19.6\%) neither (68.4\%)\\ 
				\hline
				WZ-S.amt					  & 6,579							  & racism (1.9\%) sexism (16.3\%) neither (81.8\%) \\ 
				\hline 
				WZ-S.exp					  & 6,559							  & racism (1.3\%) sexism (11.8\%) neither (86.9\%) \\ 
				\hline 
				WZ-S.gb					  	& 6,567						  			& racism (1.4\%) sexism (13.9\%) neither (84.7\%)  \\ 
				\hline 
				WZ.pj   				  	& 18,593						  & racism (10.8\%) sexism (20.3\%) neither (68.9\%)  \\ 
				\hline
				DT							  	& 24,783						  & hate (5.8\%) non-hate (94.2\%) \\ 
				\hline
				RM							  	& 2,435						  & hate (17\%) non-hate (83\%) \\ 
				\hline
    \end{tabular}
		\caption{Statistics of datasets used in the experiment}
		\label{tab_datasets}
\end{table}

\subsection{Dataset Analysis} \label{analysis}
As shown in Table \ref{tab_datasets}, all datasets are significantly biased towards non-hate, as hate tweets account between only 5.8\% (DT) and 31.6\% (WZ). When we inspect specific types of hate, some can be even more scarce, such as `racism' and as mentioned before, the extreme case of `both'. This has two implications. First, an evaluation measure such as the micro F1 that looks at a system's performance on the entire dataset regardless of class difference can be biased to the system's ability of detecting `non-hate'. In other words, a hypothetical system that achieves almost perfect F1 in identifying `racism' tweets can still be overshadowed by its poor F1 in identifying `non-hate', and vice versa. Second, compared to non-hate, the training data for hate tweets are very scarce. This may not be an issue that is easy to address as it seems, since the datasets are collected from Twitter and reflect the real nature of data imbalance in this domain. Thus to annotate more training data for hateful content we will almost certainly have to spend significantly more effort annotating non-hate. Also, as we shall show in the following, this problem may not be easily mitigated by conventional methods of over- or under-sampling. Because the real challenge is the lack of unique, discriminative linguistic characteristics in hate tweets compared to non-hate. 

As a proxy to quantify and compare the linguistic characteristics of hate and non-hate tweets, we propose to study the `uniqueness' of the vocabulary for each class. We argue that this can be a reasonable reflection of the features used for classifying each class. On the one hand, most types of features are derived from words; on the other hand, our previous work already showed that the most effective features in such tasks are based on words \cite{Robinson2018}. 

Specifically, we start with applying a state of the art tweet normalisation tool\footnote{\url{https://github.com/cbaziotis/ekphrasis}} to tokenise and transform each tweet into a sequence of words. This is done to mitigate the noise due to the colloquial nature of the data. The process involves, for example, spelling correction, elongated word normalisation (`yaaaay' becomes `yay'), word segmentation on hashtags (`\#banrefugees' becomes `ban refugees'), and unpacking contractions (e.g., `can't' becomes `can not'). Then we lemmatise each word to return its dictionary form. We refer to this process as `pre-processing' and the output as pre-processed tweets.

Next, given a tweet $t_i$, let $c_n$ be the class label of $t_i$, $words(t_i)$ returns the set of different words from $t_i$, and $uwords(c_n)$ returns the set of class-unique words that are found only for $c_n$ (i.e., they do not appear in any other classes), then for each dataset, we measure for each tweet a `uniqueness' score $u$ as:

\begin{equation} \label{e_uot}
u(t_i)= \frac{|words(t_i) \cap uwords(c_n)|} {|words(t_i)|}
\end{equation}

This measures the fraction of class-unique words in a tweet, depending on the class of this tweet. Intuitively, the score can be considered as an indication of `uniqueness' of the features found in a tweet. A high value indicates that the tweet can potentially contain more features that are unique to its class, and as a result, we can expect the tweet to be relatively easy to classify. On the other hand, a low value indicates that many features of this tweet are potentially non-discriminative as they may also be found across multiple classes, and therefore, we can expect the tweet to be relatively difficult to classify. 

\begin{figure*}
	\centering
	\includegraphics[width=430pt]{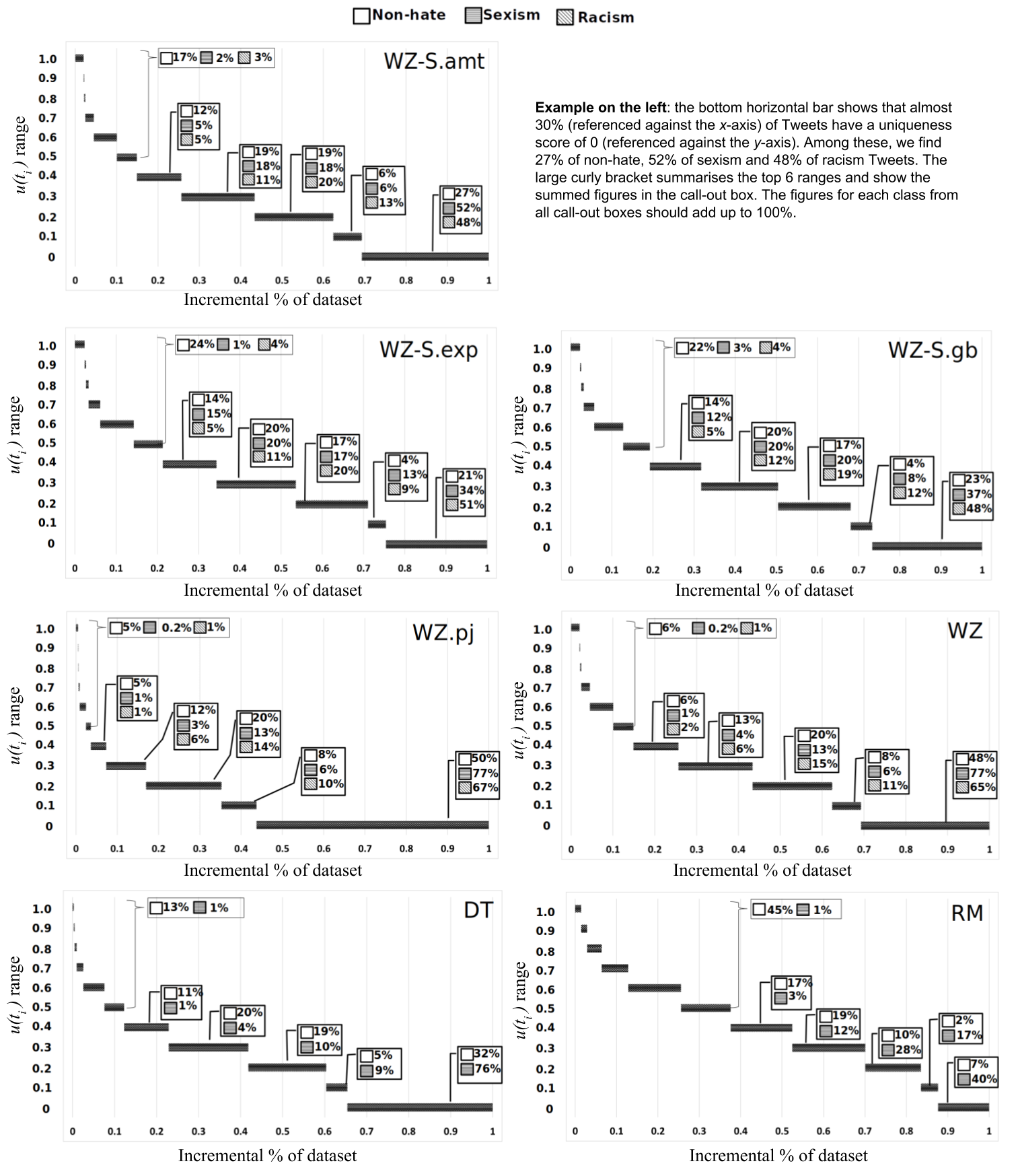}
	\caption{Distribution of tweets in each dataset over the 11 ranges of the uniqueness scores. The $x$-axis shows accumulative percentage of the dataset; the $y$-axis shows the labels of these ranges. The call-out boxes show for each class, the fraction of tweets fall under that range (in case a class is not present it has a fraction of 0\%).}    
	\label{fig_u2cit} 
\end{figure*}

We then compute this score for every tweet in a dataset, and compare the number of tweets with different uniqueness scores within each class. To better visualise this distribution, we bin the scores into 11 ranges as $u\in[0,0], u\in(0, 0.1], u\in(0.1, 0.2], ..., u\in(0.9, 1.0]$. In other words, the first range includes only tweets with a uniqueness score of 0, then the remaining 10 ranges are defined with a 0.1 increment in the score. In Figure \ref{fig_u2cit} we plot for each dataset, 1) the distribution of tweets over these ranges regardless of their class (as indicated by the length of the dark horizontal bar, measured against the $x$ axis that shows accumulative percentage of the dataset); and 2) the distribution of tweets belong to each class (as indicated by the call-out boxes). For simplicity, we label each range using its higher bound on the $y$ axis. As an example, $u\in[0,0]$ is labelled as 0, and $u\in(0,0.1]$ as 0.1. 


Using the WZ-S.amt dataset as an example, the figure shows that almost 30\% of tweets (the bottom horizontal bars in the figure) in this dataset have a uniqueness score of 0. In other words, these tweets contain no class-unique words. This can cause difficulty in extracting class-unique features from these tweets, making them very difficult to classify. The call-out box for this part of data shows that it contains 52\% of sexism and 48\% of racism tweets. In fact, on this dataset, 76\% of sexism and 81\% of racism tweets (adding up figures from the call-out boxes for the bottom three horizontal bars) only have a uniqueness score of 0.2 or lower. On those tweets that have a uniqueness score of 0.4 or higher (the top six horizontal bars), i.e., those that may be deemed as relatively `easier' to classify, we find only 2\% of sexism and 3\% of racism tweets. In contrast, it is 17\% for non-hate tweets.

We can notice very similar patterns on all the datasets in this analysis. Overall, it shows that the majority of hate tweets potentially lack discriminative features and as a result, they `sit in the long tail' of the dataset as ranked by the uniqueness of tweets. Note also that comparing the larger datasets WZ.pj and WZ against the smaller ones (i.e., the WZ-S ones), although both the absolute number and the percentage of the racism and sexism tweets are increased significantly in the two larger datasets (see Table \ref{dataset}), this does not improve the long tail situation. Indeed, one can hate or not using the same words. And as a result, increasing the dataset size and improving class balance may not always guarantee a solution.


\section{Methodology}\label{methodology}
\begin{sloppypar}
In this section, we describe our DNN based methods that implement the intuition of extracting dependency between words or phrases as features from tweets. To illustrate this idea, consider the example tweet \texttt{`These muslim refugees are troublemakers and parasites, they should be deported from my country'.} Each of the words such as `muslim', `refugee', `troublemakers', `parasites', and `deported' alone are not always indicative features of hate speech, as they can be used in any context. However, combinations such as `muslim refugees, troublemakers', `refugees, troublemakers', `refugees, parasites', `refugees, deported', and `they, deported' can be more indicative features. Clearly, in these examples, the pair of words or phrases form certain dependence on each other, and such sequences cannot be captured by n-gram like features. 
\end{sloppypar}

We propose two DNN structures that may capture such features. Our previous work \cite{Zhang2018} combines traditional a CNN with a GRU layer \cite{Zhang2018} and for the sake of completeness, we also include its details below. Our other method combines traditional CNN with some modified CNN layers serving as skip-gram extractors - to be called `skipped CNN'. Both structures modify a common, base CNN model (Section \ref{base_cnn}) that acts as the n-gram feature extractor, while the added GRU (Section \ref{cnn_gru}) and the skipped CNN (Section \ref{skipped_cnn}) components are expected to extract the dependent sequences of such n-grams, as illustrated above.

\subsection{The Base CNN Model} \label{base_cnn}
\begin{figure}
	\centering
	\includegraphics[width=180pt]{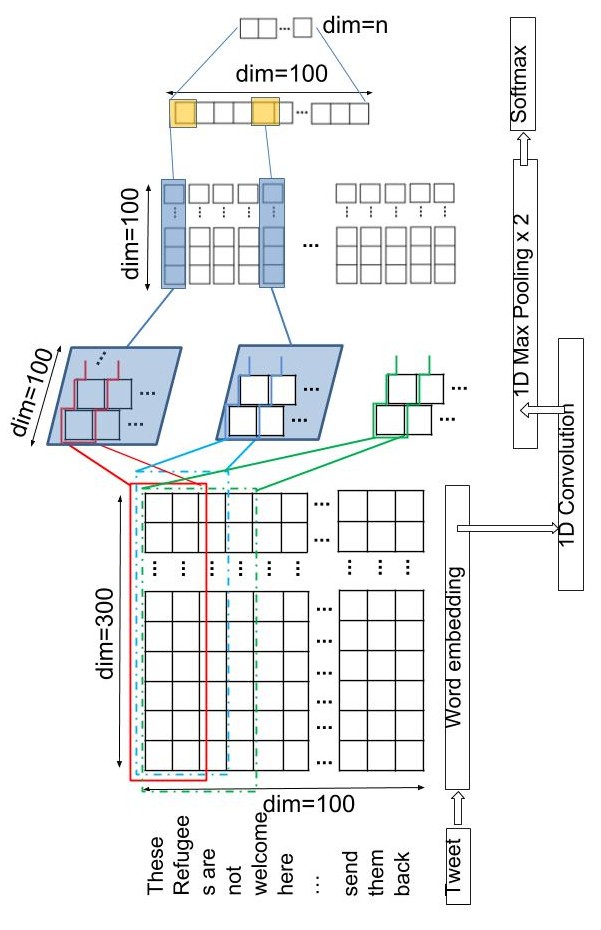}
	\caption{The Base CNN model uses three different window sizes to extract features. This diagram is best viewed in colour.}    
	\label{fig_base_cnn} 
\end{figure}

The Base CNN model is illustrated in Figure \ref{fig_base_cnn}. Given a tweet, we firstly apply the pre-processing described in Section \ref{analysis} to normalise and transform the tweet into a sequence of words. This sequence is then passed to a word embedding layer, which maps the sequence into a real vector domain (word embeddings). Specifically, each word is mapped onto a fixed dimensional real valued vector, where each element is the weight for that dimension for that word. Word embeddings are often trained on very large unlabelled corpus, and comparatively, the datasets used in this study are much smaller. Therefore in this work, we use pre-trained word embeddings that are publicly available (to be detailed in Section \ref{embedding}). One potential issue with pre-trained embeddings is Out-Of-Vocabulary (OOV) words, particularly on Twitter data due to its colloquial nature. Thus the pre-processing also helps to reduce the noise in the language and hence the scale of OOV. For example, by hashtag segmentation we transform an OOV `\#BanIslam' into `Ban' and `Islam' that are more likely to be included in the pre-trained embedding models.

The embedding layer passes an input feature space with a shape of $100 \times 300$ to three 1D convolutional layers, each uses 100 filters and a stride of 1, but different window sizes of 2, 3, and 4 respectively. Intuitively, each CNN layer can be considered as extractors of bi-gram, tri-gram and quad-gram features. The rectified linear unit function is used for activation in these CNNs. The output of each CNN is then further down-sampled by a 1D max pooling layer with a pool size of 4 and a stride of 4 for further feature selection. Outputs from the pooling layers are then concatenated, to which we add another 1D max pooling layer with the same configuration before (thus `max pooling x 2' in the figure). This is because we empirically found that this further pooling layer can lead to an improvement in F1 in most cases (with as much as 5 percentage points). The output is then fed into the final softmax layer to predict probability distribution over all possible classes ($n$), which will depend on individual datasets. 

One of the recent trends in text processing tasks on Twitter is the use of character based n-grams and embeddings instead of word based, such as in \cite{Mehdad2016,Park2017}. The main reason for this is to cope with the noisy and informal nature of the language in tweets. We do not use character based models, mainly because the literatures that compared word based and character based models are rather inconclusive. Although Mehdad et al. \cite{Mehdad2016} obtained better results using character based models, Park et al. \cite{Park2017} and Gamback et al. \cite{Gamback2017} reported the opposite. Further, our pre-processing already reduces the noise in the language to some extent. Although the state of the art tool we used is non-perfect and still made mistakes such as parsing `\#YouTube' to `You' and `Tube', overall it significantly reduced OOVs by the embedding models. Using the DT dataset for example, this improved hashtag coverage from as low as less than 1\% to up to 80\% depending on the embedding models used (see the Appendix for details). Also word-based models also better fit our intuitions explained before.


\subsubsection{CNN + GRU} \label{cnn_gru}
\begin{figure}
	\centering
	\includegraphics[width=180pt]{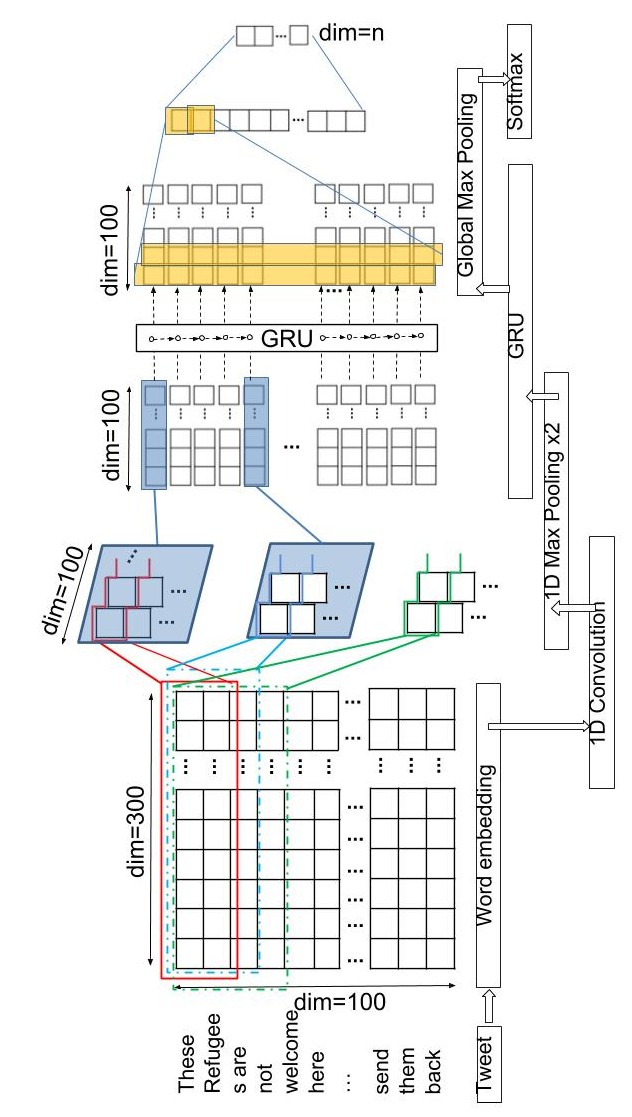}
	\caption{The CNN+GRU architecture. This diagram is best viewed in colour.}    
	\label{fig_base_gru} 
\end{figure}

With this model, we extend the Base CNN model by adding a GRU layer that takes input from the max pooling layer. This treats the features as timesteps and outputs 100 hidden units per timestep. Compared to LSTM, which is a popular type of RNN, the key difference in a GRU is that it has two gates (reset and update gates) whereas an LSTM has three gates (namely input, output and forget gates). Thus GRU is a simpler structure with fewer parameters to train. In theory, this makes it faster to train and generalise better on small data; while empirically it is shown to achieve comparable results to LSTM \cite{Chung2014}. Next, a global max pooling layer `flattens' the output space by taking the highest value in each timestep dimension, producing a feature vector that is finally fed into the softmax layer. The intuition is to pick the highest scoring features to represent a tweet, which empirically works better than the normal configuration. The structure of this model is shown in Figure \ref{fig_base_gru}.

The GRU layer captures sequence orders that can be useful for this task. In an analogy, it learns dependency relationships between n-grams extracted by the CNN layer before. And as a result, it may capture co-occurring word n-grams as useful patterns for classification, such as the pairs of words and phrases illustrated before.
	

\subsubsection{CNN + skipped CNN (sCNN)} \label{skipped_cnn}
\begin{figure}
	\centering
	\includegraphics[width=160pt]{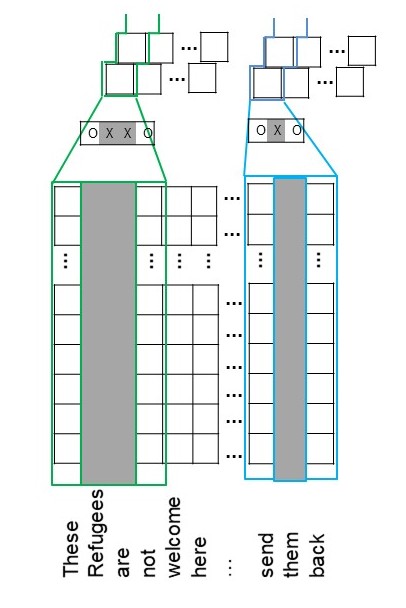}
	\caption{Example of a 2 gapped size 4 window and a one gapped size 3 window. The `X' indicates that input for the corresponding position in the window is ignored.}    
	\label{fig_gapped_window} 
\end{figure}

With this model, we propose to extend the Base CNN model by adding CNNs that use `gapped window' to extract features from its input, and we call these CNN layers `skipped CNNs'. A gapped window is one where inputs at certain (consecutive) positions of the window are ignored, such as those shown in Figure \ref{fig_gapped_window}. We say that these positions within the window are `deactivated' while other positions are `activated'. Specifically, given a window of size $j$, applying a gap of $i$ consecutive positions will produce multiple shapes of size $j$ windows, as illustrated in Algorithm \ref{alg_gapped_window}. 

\begin{algorithm}
	\caption{Creation of $i$ gapped size $j$ windows. A sequence $[$O,X,O,O$]$ represents one possible shape of a 1 gapped size 4 window, where the first and the last two positions are activated (`O') and the second position is deactivated (`X').}\label{alg_gapped_window}
	\begin{algorithmic}[1] 
		\STATE Input: $i: 0<i<j$, $j:j>0$, $w \leftarrow [p_1, ..., p_j]$
		\STATE Output: $W \leftarrow \emptyset$ a set of $j$ sized window shapes
		\FORALL{$k \in [2,j)$ and $k \in \mathbb N_{+}$}
			\STATE Set $p_1$ in $w$ to O
			\STATE Set $p_j$ in $w$ to O		
			\FORALL{$x \in [k,k+i]$ and $x \in \mathbb N_{+}$} 
				\STATE Set $p_x$ to X
				\FORALL{$y \in [k+i+1,j)$ and $y \in \mathbb N_{+}$}
					\STATE Set $p_y$ in $w$ to O			
				\ENDFOR				
				\STATE $W \leftarrow W \cup \{w\}$
			\ENDFOR
		\ENDFOR
	\end{algorithmic}
\end{algorithm}

As an example, applying a 1-gap to a size 4 window will produce two shapes: $[$O,X,O,O$]$, $[$O,O,X,O$]$, where `O' indicates an activated position and `X' indicates a deactivated position in the window; while applying a 2-gap to a size 4 window will produce a single shape of $[$O,X,X,O$]$. 

\begin{figure}
	\centering
	\includegraphics[width=220pt]{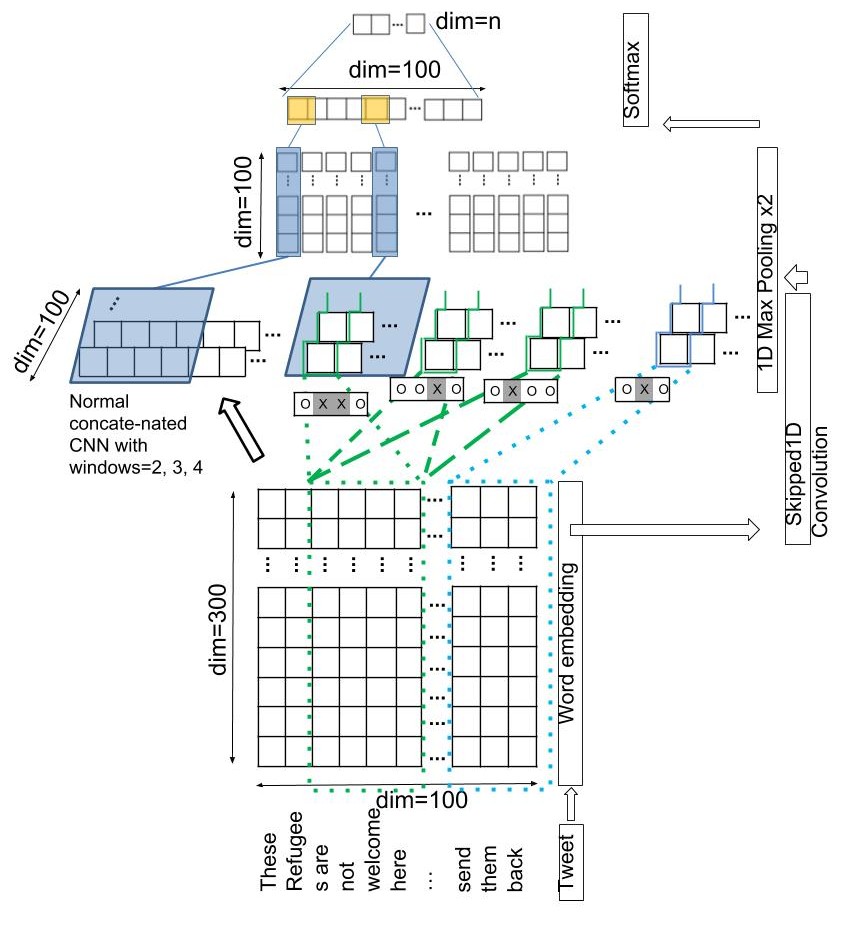}
	\caption{The CNN+sCNN model concatenates features extracted by the normal CNN layers with window sizes of 2, 3, and 4, with features extracted by the four skipped CNN layers. This diagram is best viewed in colour.}    
	\label{fig_skipped_cnn} 
\end{figure}

To extend the Base CNN model, we add CNNs using 1-gapped size 3 windows, 1-gapped size 4 windows and 2-gapped size 4 windows. Then each added CNN is followed by a max pooling layer of the same configuration as described before. The remaining parts of the structure remain the same. This results in a model illustrated in Figure \ref{fig_skipped_cnn}. 

Intuitively, the skipped CNNs can be considered as extractors of `skip-gram' like features. In an analogy, we expect it to extract useful features such as `muslim refugees ? troublemakers', `muslim ? ? troublemakers', `refugees ? troublemakers', and `they ? ? deported' from the example sentence before, where `?' is a wildcard representing any word token in a sequence. 

To the best of our knowledge, the work by Nguyen et al. \cite{NguyenG16} is the only one that uses DNN models to extract skip-gram features that are used directly in NLP tasks. However, our method is different in two ways. First, Nguyen et al. addressed a task of mention detection from sentences, i.e., classifying word tokens in a sentence into sequences of particular entities or not. Our work deals with sentence classification. This means that our modelling of the task input and their features are essentially different. Second, the authors used skip-gram features only, while our method adds skip-grams to conventional n-grams, as we concatenate the output from the skipped CNNs and the conventional CNNs. 
The concept of skip-grams has been widely quoted in training word embeddings with neural network models since Mikolov et al. \cite{Mikolov2013}. This is however, different from directly using skip-gram as features for NLP. Work such as \cite{Reyes2013} used skip-grams in detecting irony in language. But these are extracted as features in a separate process, while our method relies on the DNN structure to learn such complex features. A similar concept of atrous (or `dilated') convolution has been used in image processing \cite{Chen2016}. In comparison, given a window size of $n$ this effectively places an equal number of gaps between every element in the window. For example, a window of size 3 with a dilation rate of 2 would effectively create a window of the shape [X,O,X,O,X,O,X]. 
\newline
\newline
\noindent For both CNN+GRU and CNN+sCNN, the input to the each convolutional layer is also regularised by a dropout layer with a ratio of 0.2.

\subsubsection{Model Parameters} 
We use the categorical cross entropy loss function and the Adam optimiser to train the models, as the first is empirically found to be more effective on classification tasks than other commonly used loss functions including classification error and mean squared error \cite{McCaffrey2017}, and the second is designed to improve the classic stochastic gradient descent (SGD) optimiser and in theory combines the advantages of two other common extensions of SGD (AdaGrad and RMSProp) \cite{Kingma2015}.

Our choice of parameters described above are largely based on empirical findings reported previously, default values or anecdotal evidence. Arguably, these may not be the best settings for optimal results, which are always data-dependent. However, we show later in experiments that the models already obtain promising results even without extensive data-driven parameter tuning.

\section{Experiment}\label{experiment}
In this section, we present our experiments for evaluation and discuss the results. We compare our CNN+GRU and CNN+sCNN methods against three re-implemented state of the art methods (Section \ref{baseline}), and discuss the results in Section \ref{general_results}. This is followed by an analysis to show how our methods have managed to effectively capture hate tweets in the long tail (Section \ref{longtail_results}), and to discover the typical errors made by all methods compared (Section \ref{error_analysis}).
\newline
\newline
\noindent\textbf{Word embeddings.} We experimented with three different choices of pre-trained word embeddings: the Word2Vec embeddings trained on the 3-billion-word Google News corpus with a skip-gram model\footnote{https://github.com/mmihaltz/word2vec-GoogleNews-vectors}, the `GloVe' embeddings trained on a corpus of 840 billion tokens using a Web crawler \cite{Pennington2014}, and the Twitter embeddings trained on 200 million tweets with spam removed \cite{Li17}\footnote{`Set1' in \cite{Li17}}. However, our results did not find any embeddings that can consistently outperform others on all tasks and datasets. In fact, this is rather unsurprising, as previous studies \cite{Chiu16} suggested similar patterns: the superiority of one word embeddings model on intrinsic tasks (e.g., measuring similarity) is generally non-transferable to downstream applications, across tasks, domains, or even datasets. Below we choose to only focus on results obtained using the Word2Vec embeddings, for two reasons. On the one hand, this is consistent with previous work such as \cite{Park2017}. On the other hand, the two state of the art methods also perform best overall with Word2Vec\footnote{Based on results in Table \ref{tab_full_results_baselines} in the Appendix, on a per-class basis, Gamback et al. method obtained the highest F1 with Word2Vec embeddings on 11 out of 19 cases, with the other 8 cases obtained with either GloVe or the Twitter embeddings. For Park et al. the figure is 12 out of 19.}. Our full results obtained with the different embeddings are available in the Appendix.
\newline
\newline
\noindent\textbf{Performance evaluation metrics.} We use the standard Precision (P), Recall (R) and F1 measures for evaluation. For the sake of readability, we only present F1 scores in the following sections unless otherwise stated. Again full results can be found in the Appendix. Due to the significant class imbalance in the data, we show F1 obtained on both hate and non-hate tweets separately. 
\newline
\newline
\noindent\textbf{Implementation.} For all methods discussed in this work, we used the Python Keras\footnote{\url{https://keras.io/}, version 2.0.2} with Theano backend \footnote{\url{http://deeplearning.net/software/theano/}, version 0.9.0} and the scikit-learn\footnote{\url{http://scikit-learn.org/}, version 0.19.1} library for implementation\footnote{Code available at \url{https://github.com/ziqizhang/chase}}. For DNN based methods, we fix the \textit{epochs} to 10 and use a \textit{mini-batch} of 100 on all datasets. These parameters are rather arbitrary and fixed for consistency. We run all experiments in a 5-fold cross validation setting and report the average across all folds.

\subsection{State of the art} \label{baseline}
We re-implemented three state of the art methods covering both the classic and deep learning based methods. \textbf{First}, we use the SVM based method described in Davidson et al. \cite{Davidson2017}. A number of different types of features are used as below. Unless otherwise stated, these features are extracted from the pre-processed tweets:

\begin{itemize}
  \item Surface features: word unigram, bigram and trigram each weighted by Term Frequency Inverse Document Frequency (TF-IDF); number of mentions, and hashtags\footnote{Extracted from the original tweet before pre-processing.}; number of characters, and words;
  \item Linguistic features: Part-of-Speech (PoS)\footnote{The NLTK library is used.} tag unigrams, bigrams, and trigrams, weighted by their TF-IDF and removing any candidates with a document frequency lower than 5; number of syllables; Flesch-Kincaid Grade Level and Flesch Reading Ease scores that to measure the `readability' of a document
	\item Sentiment features: sentiment polarity scores of the tweet, calculated using a public API\footnote{\url{https://github.com/cjhutto/vaderSentiment}}. 
\end{itemize}	

Our \textbf{second} and \textbf{third} methods are re-implementations of Gamback et al. \cite{Gamback2017} (\textbf{GB}) and Park et al. \cite{Park2017} (\textbf{PK}) using word embeddings, which obtained better results than their character based counterparts in their original work. Both are based on concatenation of multiple CNNs and in fact, have the same structure as the base CNN model described in Section \ref{base_cnn} but use different CNN window sizes. They are therefore good reference for comparison to analyse the benefits of our added GRU and skipped CNN layers. We kept the same hyper-parameters for these as well as our proposed methods for comparative analysis\footnote{In fact both papers did not detail their hyper-parameter settings, which is another reason for us to use consistent configurations as our methods.}.


\subsection{Overall Results}\label{general_results}
Our experiments could not identify a best performing candidate among the three state of the art methods on all datasets, by all measures. Therefore, in the following discussion, unless otherwise stated, we compare our methods against \textit{the best results achieved by any} of the three state of the art methods
\newline
\newline
\noindent{\textbf{Overall micro and macro F1.}}
Firstly, we compare micro and macro F1 obtained by each method in Table \ref{tab_results}.  In terms of micro F1, both our CNN+sCNN and CNN+GRU methods obtained consistently the best results on all datasets. Nevertheless, the improvement over the best results from any of the three state of the art methods is rather incremental. The situation with macro F1 is however, quite different. Again compared against the state of the art, both our methods obtained consistently better results. The improvements in many cases are much higher: 1$\sim$5 and 1$\sim$4 percentage points (or `percent' in short) for CNN+sCNN and CNN+GRU respectively. When only the categories of hate tweets are considered (i.e., excluding non-hate, see `macro F1, hate'), the improvements are significant in some cases: a maximum of 8 and 6 percent for CNN+sCNN and CNN+GRU respectively. 

\begin{table*}[thb]
    \centering
    \caption{Micro v.s. macro F1 results of different methods (using the Word2Vec embeddings). The best results on each row are highlighted in \textbf{bold}. Numbers within (brackets) indicate the improvement in F1 compared to the \textit{best result} by any of the three state of the art methods.}
    \begin{tabular}{c | l | c c c| c c c c}
        \hline
        	\textbf{Dataset} &\textbf{Measure}& \textbf{SVM}	& \textbf{GB}& 
        	\textbf{PK}	&\multicolumn{2}{c}{\textbf{CNN+sCNN}}	&\multicolumn{2}{c}{\textbf{CNN+GRU}}\\\hline        
        	
        	\multirow{3}{*}{WZ-S.amt}
        			&micro F1 	&0.81 &\textbf{0.92} &\textbf{0.92} &\textbf{0.92} &(-) &\textbf{0.92} &(-)\\
        			&macro F1 	&0.59 &0.66 &0.64 &\textbf{0.68} &(+0.02) &0.67 &(+0.02)\\        			
        			&macro F1, hate 	&0.45 &0.51 &0.48 &\textbf{0.55} &(+0.04) &0.53 &(+0.02) \\\hline
    		
    		\multirow{3}{*}{WZ-S.exp}
    		&micro F1 	&0.82 &0.91 &0.91 &\textbf{0.92} &(+0.01) &\textbf{0.92} &(+0.01) \\
    		&macro F1 	&0.58 &0.70 &0.69  &\textbf{0.74} &(+0.04) &0.71 &(+0.01) \\    		
    		&macro F1, hate &0.42 &0.58 &0.56  &\textbf{0.63} &(+0.05) &0.59 &(+0.01) \\\hline
			
			\multirow{3}{*}{WZ-S.gb}
			&micro F1 &0.83 &0.92 &0.92 &\textbf{0.93} &(+0.01) &\textbf{0.93} &(+0.01) \\
			&macro F1 &0.64 &0.71 &0.70 &\textbf{0.76} &(+0.05) &0.74 &(+0.04) \\			
			&macro F1, hate 	&0.51 &0.58 &0.57 &\textbf{0.66} &(+0.08) &0.64 &(+0.06) \\\hline
			
			\multirow{3}{*}{WZ.pj}
			&micro F1 &0.72 &0.82 &0.82 &\textbf{0.83} &(+0.01) &\textbf{0.82} &(-) \\
			&macro F1 &0.66 &0.75 &0.75 &\textbf{0.77} &(+0.02) &0.76 &(+0.01) \\			
			&macro F1, hate 	&0.59 &0.68 &0.68 &\textbf{0.71} &(+0.03) &0.70 &(+0.02) \\\hline
			
			\multirow{3}{*}{WZ}
			&micro F1 &0.72 &0.82 &0.82 &\textbf{0.83} &(+0.01) &\textbf{0.82} &(-) \\
			&macro F1 &0.65 &0.75 &0.76 &\textbf{0.77} &(+0.01) &0.76 &(-) \\			
			&macro F1, hate &0.58 &0.69 &0.70 &\textbf{0.71} &(+0.01) &\textbf{0.70} &(-) \\\hline
			
			\multirow{3}{*}{DT}
			&micro F1   &0.79 &\textbf{0.94} &\textbf{0.94} &\textbf{0.94} &(-) &\textbf{0.94} &(-) \\					
			&macro F1   &0.56 &0.69 &0.63 &\textbf{0.64} &(+0.01) &0.63 &(-) \\
			&macro F1, hate 	&0.23 &0.28 &0.28 &\textbf{0.30} &(+0.02) &0.30 &(+0.02) \\\hline
			
			\multirow{3}{*}{RM}
			&micro F1   &0.79 &0.90  &0.89 &\textbf{0.91} &(+0.01) &0.90 &(-) \\					
			&macro F1 	&0.72 &0.80  &0.80 &\textbf{0.83} &(+0.03) &0.81 &(+0.01) \\
			&macro F1, hate 	&0.58 &0.67 &.66 &\textbf{0.71} &(+0.04) &0.68 &(+0.02) \\
				\hline

    \end{tabular}
		
		\label{tab_results}
\end{table*}


However, notice that both the overall and hate speech-only macro F1 scores are significantly lower than micro F1, for any methods, on any datasets. This further supports our earlier data analysis findings that classifying hate tweets is a much harder task than non-hate, and micro F1 scores will overshadow a method's true performance on a per-class basis, due to the imbalanced nature of such data. 
\newline
\newline
\noindent{\textbf{F1 per-class.}} To clearly compare each method's performance on classifying hate speech, we show the per-class results in Table \ref{tab_results_per_class}. This demonstrates the benefits of our methods compared to the state of the art when focusing on any categories of hate tweets rather than non-hate: the CNN+sCNN model was able to outperform the best results by any of the three comparison methods, achieving a maximum of 13 percent increase in F1; the CNN+GRU model was less good, but still obtained better results on four datasets, achieving a maximum of 8 percent increase in F1. 

\begin{table*}[thb]
    \centering
    \caption{F1 results of different models for each class (using the Word2Vec embeddings). The best results on each row are highlighted in \textbf{bold}. Numbers within (brackets) indicate the improvement in F1 compared to the \textit{best results} from any of the three state of the art methods.}
    \begin{tabular}{c | c | c c c| c c c c}
        \hline
        	\textbf{Dataset} &\textbf{Class}& \textbf{SVM}	& \textbf{GB}& 
        	\textbf{PK}	&\multicolumn{2}{c}{\textbf{CNN+sCNN}}	&\multicolumn{2}{c}{\textbf{CNN+GRU}}\\\hline        
        	
        	\multirow{3}{*}{WZ-S.amt}
        			&racism 	&0.22 &0.22 &0.17 &\textbf{0.29} &(+0.07) &0.28 &(+0.06)\\
        			&sexism 	&0.68 &0.80 &0.80 &\textbf{0.81} &(+0.01) &0.80 &(-)\\        			
        			&non-hate 	&0.88 &\textbf{0.95} &\textbf{0.95} &\textbf{0.95} &(-) &\textbf{0.95} &(-) \\\hline
    		
    		\multirow{3}{*}{WZ-S.exp}
    		&racism 	&0.26 &0.51 &0.45  &\textbf{0.58} &(+0.07) &0.51 &(-) \\
    		&sexism 	&0.58 &0.66 &0.67  &\textbf{0.68} &(+0.01) &0.66 &(-) \\    		
    		&non-hate   &0.89 &0.95 &0.95  &\textbf{0.96} &(+0.01) &0.95 &(-) \\\hline
			
			\multirow{3}{*}{WZ-S.gb}
			&racism 	&0.38 &0.41 &0.38 &\textbf{0.54} &(+0.13) &0.49 &(+0.08) \\
			&sexism 	&0.64 &0.76 &0.76 &\textbf{0.77} &(+0.02) &0.78 &(+0.02) \\			
			&non-hate 	&0.90 &\textbf{0.96} &\textbf{0.96} &\textbf{0.96} &(-) &\textbf{0.96} &(-) \\\hline
			
			\multirow{3}{*}{WZ.pj}
			&racism 	&0.60 &0.70 &0.69 &\textbf{0.73} &(+0.03) &\textbf{0.73} &(+0.03) \\
			&sexism 	&0.58 &0.66 &0.67 &\textbf{0.69} &(+0.02) &0.68 &(+0.01) \\			
			&non-hate 	&0.80 &\textbf{0.88} &\textbf{0.88} &\textbf{0.88} &(-) &\textbf{0.87} &(-0.01) \\\hline
			
			\multirow{3}{*}{WZ}
			&racism 	&0.59 &0.70 &0.72 &\textbf{0.74} &(+0.02) &0.73 &(+0.01) \\
			&sexism 	&0.57 &0.66 &0.66 &\textbf{0.68} &(+0.02) &\textbf{0.67} &(+0.01) \\			
			&non-hate   &0.79 &0.87 &\textbf{0.88} &\textbf{0.88} &(+0.01) &\textbf{0.87} &(-0.01) \\\hline
			
			\multirow{2}{*}{DT}
			&hate 	    &0.23 &0.28 &0.28 &\textbf{0.30} &(+0.02) &0.29 &(+0.01) \\					
			&non-hate 	&0.88 &\textbf{0.97} &\textbf{0.97} &\textbf{0.97} &(-) &\textbf{0.97} &(-) \\\hline
			
			\multirow{2}{*}{RM}
			&hate 	    &0.58 &0.67  &0.66 &\textbf{0.71} &(+0.04) &0.68 &(+0.01) \\					
			&non-hate 	&0.86 &\textbf{0.94}  &\textbf{0.94} &\textbf{0.94} &(-) &\textbf{0.94} &(-) \\
				\hline

    \end{tabular}
		
		\label{tab_results_per_class}
\end{table*}


Note that the best improvement obtained by our methods were found on the racism class in the three WZ-S datasets. As shown in Table \ref{tab_datasets}, these are minority classes representing a very small population in the dataset (between 1 and 6\%). This suggests that our proposed methods can be potentially very effective when there is a lack of training data. 

It is also worth to highlight that here we discuss only results based on the Word2Vec embeddings. In fact, our methods obtained even more significant improvement when using the Twitter or GloVe embeddings. We discuss this further in the following sections.
\newline
\newline
\begin{sloppypar}
\noindent{\textbf{sCNN v.s. GRU.}} Comparing CNN+sCNN with CNN+GRU using Table \ref{tab_results_per_class}, we see that the first performed much better when only hate tweets are considered, suggesting that the skipped CNNs may be more effective feature extractors than GRU for hate speech detection in very short texts such as tweets. CNN+sCNN consistently outperformed the highest F1 by any of the three state of the art methods on any dataset, and it also achieved higher improvement. CNN+GRU on the other hand, obtained better F1 on four datasets and the same best F1 (as the state of the art) on two datasets. This also translates to overall better macro F1 by CNN+sCNN compared to CNN+GRU (Table \ref{tab_results}). 
\newline
\newline
\noindent{\textbf{Patterns observed with other word embeddings.}}
While we show detailed results with different word embeddings in the Appendix, it is worth to mention here that the same patterns were noticed with these different embeddings, i.e., both CNN+sCNN and CNN+GRU have outperformed the state of the art methods in the most cases but the first model obtained much better results. The best results however, were not always obtained with the Word2Vec embeddings on all datasets. On an individual class basis, among all the 19 classes from the seven datasets, the CNN+sCNN model has scored 10 best F1 with the Word2Vec embeddings, 11 best F1 with the Twitter embeddings, and 10 best F1 with the GloVe embeddings. For the CNN+GRU model, the figures are 14, 12, and 9 for the Word2Vec, Twitter, and GloVe embeddings respectively. 

Both models however, obtained much more significant improvement over the three state of the art methods when using the Twitter or GloVe embeddings instead of Word2Vec. For example, on the three WZ-S datasets with the smallest class `racism', CNN+sCNN outperformed the best results by the three state of the art by between 20 and 22 percentage points in F1 using the Twitter embeddings, or between 21 and 33 percent using the GloVe embeddings. For CNN+GRU, the situation is similar: between 9 and 18 percent using the Twitter embeddings, or between 11 and 20 percent using the GloVe embeddings. However, this is largely because the GB and PK models under-performed significantly when using these embeddings instead of Word2Vec. In contrast, the CNN+sCNN and CNN+GRU models can be seen to be less sensitive to the choice of embeddings, which can be a desirable feature.
\newline
\newline
\noindent{\textbf{Against previously reported results.}}
For many reasons such as the difference in the re-generated datasets\footnote{As noted in \cite{Zhang2018}, we had to re-download tweets using previously published tweet IDs in the shared datasets. But some tweets have become no longer available.}, possibly different pre-processing methods, and unknown hyper-parameter settings from the previous work\footnote{The details of the pre-processing, network structures and many hyper-parameter settings are not reported in nearly all of the previous work. For comparability, as mentioned before, we kept the same configurations for our methods as well as the re-implemented state of the art methods.}, we cannot guarantee an identical replication of the previous methods in our re-implementation. Therefore in Table \ref{tab_results_soa}, we compare results by our methods against previously reported results (micro F1 is used as this is the case in all the previous work) on each dataset on an `as-is' basis. 

\begin{table}[thb]
    \centering
    \caption{Comparing micro F1 on each dataset (using the Word2Vec embeddings) against previously reported results on an `as-is' basis. The best performing result on each dataset is highlighted in \textbf{bold}. For \cite{Waseem2016a} and \cite{Waseem2016b}, we used the result reported under their `Best Feature' setting. }
    \begin{tabular}{p{1.2cm}|p{0.5cm} p{0.4cm}|p{1.2cm}p{1.2cm}}
        \hline
        	Dataset&\multicolumn{2}{c|}{State of the art}	& CNN+sCNN	& CNN+GRU\\\hline                
    
			WZ-S.amt&0.84 		&\cite{Waseem2016b}
													&\textbf{0.92}  &\textbf{0.92}  \\
			
			WZ-S.exp&0.91 		&\cite{Waseem2016b}
			&\textbf{0.92}  &\textbf{0.92}  \\													
						
			WZ-S.gb&0.78 		&\cite{Gamback2017} 
			&\textbf{0.93}  &\textbf{0.93}  \\													

			WZ.pj  &\textbf{0.83}	&\cite{Park2017} 
			&\textbf{0.83}  &\textbf{0.83}  \\	
			
			WZ  	&0.74	&\cite{Waseem2016a} 
			&\textbf{0.83}  &\textbf{0.83}  \\	

			DT  	&0.90	&\cite{Davidson2017} 
			&\textbf{0.94}  &\textbf{0.94}  \\	
			
				\hline

    \end{tabular}
		
		\label{tab_results_soa}
\end{table}


Table \ref{tab_results_soa} shows that both our methods have achieved the best results on all datasets, outperforming state of the art on six and in some cases, quite significantly. Note that on the WZ.pj dataset where our methods did not obtain further improvement, the best reported state of the art result was obtained using a hybrid character-and-word embeddings CNN model \cite{Park2017}. Our methods in fact, outperformed both the word-only and character-only embeddings models in that same work.
\end{sloppypar}

\subsection{Effectiveness on Identifying the Long Tail}\label{longtail_results}
While the results so far have shown that our proposed methods can obtain better performance in the task, it is unclear whether they are particularly effective on classifying tweets in the long tail of such datasets as we have shown before in Figure \ref{fig_u2cit}. To understand this, we undertake a further analysis below.

\begin{figure}
	\centering
	\includegraphics[width=220pt]{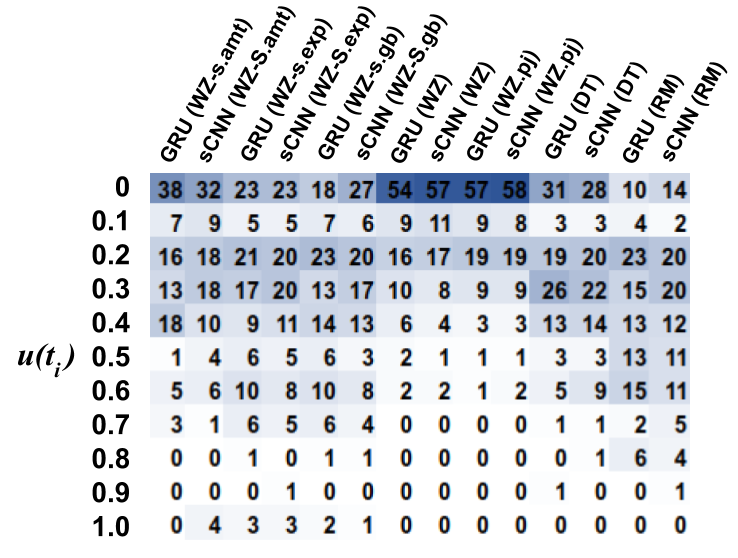}
	\caption{(Best viewed in colour) Distribution of additional true positives (compared against Gamback et al.) identified by CNN+sCNN (sCNN for shorthand) and CNN+GRU (GRU) over different ranges of uniqueness scores (Equation \ref{e_uot}) when using the Word2Vec embeddings. Each row in the heatmap corresponds to a uniqueness score range. Each column corresponds to a method-dataset pair. The numbers in each column sum up to 100\% while the colour scale within each cell is determined by the number in that cell.}    
	\label{fig_heatmap} 
\end{figure}

On each dataset, we compare the output from our proposed methods against that from Gamback et al. as a reference. We identify the additional tweets that were correctly classified by either of our CNN+sCNN or CNN+GRU methods. We refer to these tweets as \textbf{additional true positives}. Next, following the same process as that for Figure \ref{fig_u2cit}, we 1) compute the uniqueness score of each tweet (Equation \ref{e_uot}) as indicator of the fraction of class-unique words in each of them; 2) bin the scores into 11 ranges; and 3) show the distribution of additional true positives found on each dataset by our methods over these ranges in Figure \ref{fig_heatmap} (i.e., each column in the figure corresponds to a method-dataset pair). Again for simplicity, we label each range using its higher bound on the $y$ axis (e.g., $u\in[0,0]$ is labelled as 0, and $u\in(0,0.1]$ as 0.1). As an example, the leftmost column shows that comparing the output from our CNN+GRU model against Gamback et al. on the WZ-S.amt dataset, 38\% of the additional true positives have a uniqueness score of 0.

The figure shows that in general, the vast majority of additional true positives have low uniqueness scores. The pattern is particularly strong on WZ and WZ.pj datasets, where the majority of additional true positives have very low uniqueness scores and in a very large number of cases, a substantial fraction of them (between 50 and 60\%) have $u(t_i)=0$, suggesting that these tweets have no class-unique words at all and therefore, we expect them to potentially have fewer class-unique features. However, our methods still managed to classify them correctly, while the method by Gamback et al. could not. We believe these results are convincing evidence that our methods of using skipped CNN or GRU structures on such tasks can significantly improve our capability of classifying tweets that lack discriminative features. Again the results shown in Figure \ref{fig_heatmap} are based on the Word2Vec embeddings but we noticed the same patterns with other embeddings, as shown in Figure \ref{fig_heatmap_full} in the Appendix.

\subsection{Error Analysis} \label{error_analysis}
\begin{sloppypar}
To understand further the challenges of the task, we manually analysed a sample of 200 tweets covering all classes to identify ones that are incorrectly classified by all methods. We generally split these errors into four categories.
\newline
\newline
\noindent\textbf{Implicitness} (46\%) represents the largest group of errors, where the tweet does not contain explicit lexical or syntactic patterns as useful classification features. Interpretation of such tweets almost certainly requires complicated reasoning and cultural and social background knowledge. For example, subtle metaphors such as \texttt{`expecting gender equality is the same as genocide'}, stereotypical views such as in \texttt{`... these same girls ... didn't cook that well and aren't very nice'} are often found in false negative hate tweets. 
\newline
\newline
\noindent\textbf{Non-discriminative features} (24\%) is the second majority case, where the classifiers were confused by certain features that are frequent, seemingly indicative of hate speech but in fact, can be found in both hate and non-hate tweets. For example, one would assume that the presence of the phrase \texttt{`white trash'} or pattern \texttt{`* trash'} is more likely to be a strong indicator of hate speech than not, such as in \texttt{`White bus drivers are all white trash...'}. However, our analysis shows that such seemingly `obvious' features are also prevalent in non-hate tweets such as \texttt{`... I'm a piece of white trash I say it proudly'}. The second example does not qualify as hate speech since it does not `target individual or groups' or `has the intention to incite harm'. 
\newline
\newline
\noindent There is also a large group of tweets that require interpretation of \textbf{contextual information} (18\%) such as the threads of conversation that they are part of, or from the included URLs in the tweets to fully understand their nature. In these cases, the language alone often has no connotation of hate. For example, in \texttt{`what they tell you is their intention is not their intention. https://t.co/8cmfoOZwxz'}, the language itself does not imply hate. However, when it is combined with the video content referenced by the link, the tweet incites hatred towards particular religious groups. The content referenced by URLs can be videos, images, websites, or even other tweets. Another example is \texttt{`@XXX Doing nothing does require an inordinate amount of skill'} (where `XXX' is anonymised Twitter user handle) that is part of a conversation that makes derogatory remarks towards feminists.
\newline
\newline
\noindent Finally, we also identify a fair amount of \textbf{disputable annotations} (12\%) that we could not completely agree with, even if the context as discussed above has been taken into account. For example, the tweet \texttt{`@XXX @XXX He got one serve, not two. Had to defend the doubles lines also'} is part of a conversation discussing a sports event and is annotated as sexism in the WS.pj dataset. However, we did not find anything of a hateful nature in the conversation. Another example in the same dataset is \texttt{`@XXX Picwhatting? And you have quoted none of the tweets. What are you trying to say ...?'} is questioning a point raised in another tweet which we consider as sexism, but this tweet itself has been annotated as sexism.
\end{sloppypar}

Taking into all such examples into consideration, we see that completely detecting hateful tweets purely based on their linguistic content still remains extremely challenging, if not impossible. 
\section{Conclusion and Future Work}\label{conclusion}
The propagation of hate speech on social media has been increasing significantly in recent years, both due to the anonymity and mobility of such platforms, as well as the changing political climate from many places in the world. Despite substantial effort from law enforcement departments, legislative bodies as well as millions of investment from social media companies, it is widely recognised that effective counter-measures rely on automated semantic analysis of such content. A crucial task in this direction is the detection and classification of hate speech based on its targeting characteristics. 

This work makes several contributions to state of the art in this research area. Firstly, we undertook a thorough data analysis to understand the extremely unbalanced nature and the lack of discriminative features of hateful content in the typical datasets one has to deal with in such tasks. Secondly, we proposed new DNN based methods for such tasks, particularly designed to capture implicit features that are potentially useful for classification. Finally, our methods were thoroughly evaluated on the largest collection of Twitter datasets for hate speech, to show that they can be particularly effective on detecting and classifying hateful content (as opposed to non-hate), which we have shown to be more challenging and arguably more important in practice. Our results set a new benchmarking reference in this area of research.
\newline
\newline
\noindent\textbf{Lessons learned.} \textit{First}, we showed that the very challenging nature of identifying hate speech from short text such as tweets is due to the fact that hate tweets are found in the long tail of a dataset due to their lack of unique, discriminative features. We further showed in experiments that for this very reason, the practice of `micro-averaging' over both hate and non-hate classes in a dataset adopted for reporting results by most previous work can be questionable. It can significantly bias the evaluation towards the dominant non-hate class in a dataset, overshadowing a method's ability to identify real hate speech. 

\textit{Second}, our proposed `skipped' CNN or GRU structures are able to discover implicit features that can be potentially useful for identifying hate tweets in the long tail. Interestingly, this may suggest that both structures can be potentially effective in the case where there is a lack of training data, and we plan to further evaluate this in the future. Among the two, the skipped CNNs performs much better.
\newline
\newline
\noindent\textbf{Future work.} We aim to explore the following directions of research in the future. 

\textit{First}, we will explore the options to apply our concept of skipped CNNs to character embeddings, which can further address the problem of OOVs in word embeddings. A related limitation of our work is the lack of understanding of the effect of tweet normalisation on the accuracy of the classifiers. This can be a rather complex problem as our preliminary analysis showed no correlation between the size of OOVs and classifier performance. We will investigate into this further.

\textit{Second}, we will explore other branches of methods that aim at compensating the lack of training data in supervised learning tasks. Methods such as transfer learning could be potentially promising, as they study the problem of adapting supervised models trained in a resource-rich context to a resource-scare context. We will investigate, for example, whether features discovered from one hate class can be transferred to another, thus enhancing the training of each other.

\textit{Third}, as shown in our data analysis as well as error analysis, the presence of abstract concepts such as `sexism', `racism' or even `hate' in general is very difficult to detect if solely based on linguistic content. Therefore, we see the need to go beyond pure text classification and explore possibilities to model and integrate features about users, social groups, mutual communication and even background knowledge (e.g., concepts expressed from tweets) encoded in existing semantic knowledge bases. For example, recent work \cite{Ribeiro2017} has shown that user characteristics such as their vocabulary, overall sentiment in their tweets, and network status (e.g., centrality, influence) can be useful predictors for abusive content.

\textit{Finally}, our methods prove to be effective for classifying tweets, a type of short texts. We aim to investigate whether the benefits of such DNN structures can generalise to other short text classification tasks, such as in the context of sentences.

\bibliographystyle{plain}
\bibliography{chase}

\begin{thebibliography}{10}

\bibitem{Badjatiya2017}
Pinkesh Badjatiya, Shashank Gupta, Manish Gupta, and Vasudeva Varma.
\newblock Deep learning for hate speech detection in tweets.
\newblock In {\em Proceedings of the 26th International Conference on World
  Wide Web Companion}, WWW '17 Companion, pages 759--760, Republic and Canton
  of Geneva, Switzerland, 2017. International World Wide Web Conferences
  Steering Committee.
\newblock doi:10.1145/3041021.3054223.

\bibitem{Burnap2015}
Pete Burnap and Matthew~L. Williams.
\newblock Cyber hate speech on twitter: An application of machine
  classification and statistical modeling for policy and decision making.
\newblock {\em Policy and Internet}, 7(2):223--242, 2015.
\newblock doi:10.1002/poi3.85.

\bibitem{Burnap2016}
Pete Burnap and Matthew~L. Williams.
\newblock Us and them: identifying cyber hate on twitter across multiple
  protected characteristics.
\newblock {\em EPJ Data Science}, 5(11):1--15, 2016.
\newblock doi:10.1140/epjds/s13688-016-0072-6.

\bibitem{Chen2016}
Liang-Chieh Chen, George Papandreou, Iasonas Kokkinos, Kevin Murphy, and Alan~L
  Yuille.
\newblock Deep{L}ab: Semantic image segmentation with deep convolutional nets,
  atrous convolution, and fully connected {CRFs}.
\newblock {\em IEEE Transactions on Pattern Analysis and Machine Intelligence},
  40(4):834--848, 2018.
\newblock doi:10.1109/TPAMI.2017.2699184.

\bibitem{Chen2012}
Ying Chen, Yilu Zhou, Sencun Zhu, and Heng Xu.
\newblock Detecting offensive language in social media to protect adolescent
  online safety.
\newblock In {\em Proceedings of the 2012 ASE/IEEE International Conference on
  Social Computing and 2012 ASE/IEEE International Conference on Privacy,
  Security, Risk and Trust}, SOCIALCOM-PASSAT '12, pages 71--80, Washington,
  DC, USA, 2012. IEEE Computer Society.
\newblock doi:10.1109/SocialCom-PASSAT.2012.55.

\bibitem{Chiu16}
Billy Chiu, Anna Korhonen, and Sampo Pyysalo.
\newblock Intrinsic evaluation of word vectors fails to predict extrinsic
  performance.
\newblock In {\em Proceedings of the 1st Workshop on Evaluating Vector-Space
  Representations for NLP}, pages 1--6. Association for Computational
  Linguistics, 2016.
\newblock doi:10.18653/v1/W16-2501.

\bibitem{Chung2014}
Junyoung Chung, Caglar Gulcehre, KyungHyun Cho, and Yoshua Bengio.
\newblock Empirical evaluation of gated recurrent neural networks on sequence
  modeling.
\newblock In {\em Deep Learning and Representation Learning Workshop at the
  28th Conference on Neural Information Processing Systems}, New York, USA,
  2014. Curran Associates.

\bibitem{Dadvar2013}
Maral Dadvar, Dolf Trieschnigg, Roeland Ordelman, and Franciska de~Jong.
\newblock Improving cyberbullying detection with user context.
\newblock In {\em Proceedings of the 35th European Conference on Advances in
  Information Retrieval}, ECIR'13, pages 693--696, Berlin/Heidelberg, Germany,
  2013. Springer.
\newblock doi:10.1007/978-3-642-36973-5{\_}62.

\bibitem{Davidson2017}
Thoams Davidson, Dana Warmsley, Michael Macy, and Ingmar Weber.
\newblock Automated hate speech detection and the problem of offensive
  language.
\newblock In {\em Proceedings of the 11th Conference on Web and Social Media},
  Menlo Park, California, United States, 2017. Association for the Advancement
  of Artificial Intelligence.

\bibitem{Dinakar2012}
Karthik Dinakar, Birago Jones, Catherine Havasi, Henry Lieberman, and Rosalind
  Picard.
\newblock Common sense reasoning for detection, prevention, and mitigation of
  cyberbullying.
\newblock {\em ACM Transactions on Interactive Intelligent Systems},
  2(3):18:1--18:30, 2012.
\newblock doi@10.1145/2362394.2362400.

\bibitem{Djuric2015}
Nemanja Djuric, Jing Zhou, Robin Morris, Mihajlo Grbovic, Vladan Radosavljevic,
  and Narayan Bhamidipati.
\newblock Hate speech detection with comment embeddings.
\newblock In {\em Proceedings of the 24th International Conference on World
  Wide Web}, WWW '15 Companion, pages 29--30, New York, NY, USA, 2015. ACM.
\newblock doi:10.1145/2740908.2742760.

\bibitem{EEA2017}
EEANews.
\newblock Countering hate speech online, Last accessed: July 2017,
  http://eeagrants.org/News/2012/.

\bibitem{Galiardone2015}
Igini Galiardone, Danit Gal, Thiago Alves, and Gabriela Martinez.
\newblock Countering online hate speech.
\newblock {\em UNESCO Series on Internet Freedom}, pages 1--73, 2015.

\bibitem{Gamback2017}
Bj\"{o}rn Gamb\"{a}ck and Utpal~Kumar Sikdar.
\newblock Using convolutional neural networks to classify hate speech.
\newblock In {\em Proceedings of the First Workshop on Abusive Language
  Online}, pages 85--90. Association for Computational Linguistics, 2017.
\newblock doi:10.18653/v1/W17-3013.

\bibitem{Gitari2015}
Njagi~Dennis Gitari, Zhang Zuping, Hanyurwimfura Damien, and Jun Long.
\newblock A lexicon-based approach for hate speech detection.
\newblock {\em International Journal of Multimedia and Ubiquitous Engineering},
  10(10):215--230, 2015.
\newblock doi:10.14257/ijmue.2015.10.4.21.

\bibitem{Greevy2004}
Edel Greevy and Alan~F. Smeaton.
\newblock Classifying racist texts using a support vector machine.
\newblock In {\em Proceedings of the 27th Annual International ACM SIGIR
  Conference on Research and Development in Information Retrieval}, SIGIR '04,
  pages 468--469, New York, NY, USA, 2004. ACM.
\newblock doi:10.1145/1008992.1009074.

\bibitem{TheGuardian2017}
Guardian.
\newblock Anti-muslim hate crime surges after manchester and london bridge
  attacks, Last accessed: July 2017, https://www.theguardian.com.

\bibitem{TheGuardian2016}
Guardian.
\newblock Zuckerberg on refugee crisis: `hate speech has no place on
  {F}acebook', Last accessed: July 2017, https://www.theguardian.com.

\bibitem{Kingma2015}
Diederik~P. Kingma and Jimmy Ba.
\newblock Adam: A method for stochastic optimization.
\newblock In {\em Proceedings of the 3rd International Conference for Learning
  Representations}, 2015.

\bibitem{Kwok2013}
Irene Kwok and Yuzhou Wang.
\newblock Locate the hate: Detecting tweets against blacks.
\newblock In {\em Proceedings of the Twenty-Seventh AAAI Conference on
  Artificial Intelligence}, AAAI'13, pages 1621--1622, Menlo Park, California,
  United States, 2013. Association for the Advancement of Artificial
  Intelligence.

\bibitem{Li17}
Quanzhi Li, Sameena Shah, Xiaomo Liu, and Armineh Nourbakhsh.
\newblock Data sets: Word embeddings learned from tweets and general data.
\newblock In {\em Proceedings of the Eleventh International Conference on Web
  and Social Media}, pages 428--436, Menlo Park, California, United States,
  2017. Association for the Advancement of Artificial Intelligence.

\bibitem{Lomas2015}
Natasha Lomas.
\newblock {F}acebook, google, twitter commit to hate speech action in germany,
  Last accessed: July 2017.

\bibitem{McCaffrey2017}
James~D. McCaffrey.
\newblock Why you should use cross-entropy error instead of classification
  error or mean squared error for neural network classifier training, Last
  accessed: Jan 2018, https://jamesmccaffrey.wordpress.com.

\bibitem{Mehdad2016}
Yashar Mehdad and Joel Tetreault.
\newblock Do characters abuse more than words?
\newblock In {\em Proceedings of the 17th Annual Meeting of the Special
  Interest Group on Discourse and Dialogue}, pages 299--303. Association for
  Computational Linguistics, 2016.
\newblock doi:10.18653/v1/W16-3638.

\bibitem{Mikolov2013}
Tomas Mikolov, Kai Chen, Greg Corrado, and Jeffrey Dean.
\newblock Efficient estimation of word representations in vector space.
\newblock {\em CoRR}, abs/1301.3781, 2013.

\bibitem{NguyenG16}
Thien~Huu Nguyen and Ralph Grishman.
\newblock Modeling skip-grams for event detection with convolutional neural
  networks.
\newblock In {\em Proceedings of the 2016 Conference on Empirical Methods in
  Natural Language Processing}, pages 886--891. Association for Computational
  Linguistics, 2016.
\newblock doi:10.18653/v1/D16-1085.

\bibitem{Nobata2016}
Chikashi Nobata, Joel Tetreault, Achint Thomas, Yashar Mehdad, and Yi~Chang.
\newblock Abusive language detection in online user content.
\newblock In {\em Proceedings of the 25th International Conference on World
  Wide Web}, WWW '16, pages 145--153, Republic and Canton of Geneva,
  Switzerland, 2016. International World Wide Web Conferences Steering
  Committee.
\newblock do:10.1145/2872427.2883062.

\bibitem{Nockleby2000}
John~T. Nockleby.
\newblock {\em Hate Speech}, pages 1277--1279.
\newblock Macmillan, New York, 2000.

\bibitem{Okeowo2016}
A.~Okeowo.
\newblock Hate on the rise after {T}rump's election, Last accessed: July 2017,
  http://www.newyorker.com/.

\bibitem{Ordonez2016}
Francisco~Javier Ord\'{o}\~{n}ez and Daniel Roggen.
\newblock Deep convolutional and {LSTM} recurrent neural networks for
  multimodal wearable activity recognition.
\newblock {\em Sensors}, 16(1), 2016.
\newblock doi:10.3390/s16010115.

\bibitem{Park2017}
Jo~Ho Park and Pascale Fung.
\newblock One-step and two-step classification for abusive language detection
  on {T}witter.
\newblock In {\em ALW1: 1st Workshop on Abusive Language Online}, pages 41--45,
  Vancouver, Canada, 2017. Association for Computational Linguistics.
\newblock doi:10.18653/v1/W17-3006.

\bibitem{Pennington2014}
Jeffrey Pennington, Richard Socher, and Christopher~D. Manning.
\newblock Glo{V}e: Global vectors for word representation.
\newblock In {\em Empirical Methods in Natural Language Processing (EMNLP)},
  pages 1532--1543. Association for Computational Linguistics, 2014.
\newblock doi:10.3115/v1/D14-1162.

\bibitem{Reyes2013}
Antonio Reyes, Paolo Rosso, and Tony Veale.
\newblock A multidimensional approach for detecting irony in {T}witter.
\newblock {\em Language Resources and Evaluation}, 47(1):239--268, 2013.
\newblock doi:10.1007/s10579-012-9196-x.

\bibitem{Ribeiro2017}
Manoel~Horta Ribeiro, Perod Calais, Yuri Santos, Virgilio Almeida, and Wagner
  Meira.
\newblock `{L}ike sheep among wolves': {C}haracterizing hateful users on
  {T}witter.
\newblock In {\em Proceedings of the Workshop on Misinformation and Misbehavior
  Mining on the Web, at the 11th ACM International Conference on Web Search and
  Data Mining}, New York, US, 2017. ACM.

\bibitem{Robinson2018}
David Robinson, Ziqi Zhang, and John Tepper.
\newblock Hate speech detection on {T}witter: feature engineering v.s. feature
  selection.
\newblock In {\em Proceedings of the 15th Extended Semantic Web Conference,
  Poster Volume}, ESWC'18, pages 46--49, Berlin, Germany, 2018. Springer.
\newblock doi:10.1007/978-3-319-98192-5{\_}9.

\bibitem{Schmidt2017}
Anna Schmidt and Michael Wiegand.
\newblock A survey on hate speech detection using natural language processing.
\newblock In {\em International Workshop on Natural Language Processing for
  Social Media}, pages 1--10. Association for Computational Linguistics, 2017.
\newblock doi:10.18653/v1/W17-1101.

\bibitem{Vigna2017}
Fabio~Del Vigna, Andrea Cimino, Felice Dell'Orletta, Marinella Petrocchi, and
  Maurizio Tesconi.
\newblock Hate me, hate me not: Hate speech detection on {F}acebook.
\newblock In {\em Proceedings of the First Italian Conference on
  Cybersecurity}, pages 86--95. CEUR Workshop Proceedings, 2017.

\bibitem{Warner2012}
William Warner and Julia Hirschberg.
\newblock Detecting hate speech on the {World Wide Web}.
\newblock In {\em Proceedings of the Second Workshop on Language in Social
  Media}, LSM '12, pages 19--26. Association for Computational Linguistics,
  2012.

\bibitem{Waseem2016b}
Zeerak Waseem.
\newblock Are you a racist or am i seeing things? {Annotator} influence on hate
  speech detection on {Twitter}.
\newblock In {\em Proceedings of the Workshop on NLP and Computational Social
  Science}, pages 138--142. Association for Computational Linguistics, 2016.
\newblock doi:10.18653/v1/W16-5618.

\bibitem{Waseem2016a}
Zeerak Waseem and Dirk Hovy.
\newblock Hateful symbols or hateful people? {Predictive} features for hate
  speech detection on {Twitter}.
\newblock In {\em Proceedings of the NAACL Student Research Workshop}, pages
  88--93. Association for Computational Linguistics, 2016.
\newblock doi:10.18653/v1/N16-2013.

\bibitem{Xiang2012}
Guang Xiang, Bin Fan, Ling Wang, Jason Hong, and Carolyn Rose.
\newblock Detecting offensive tweets via topical feature discovery over a large
  scale {Twitter} corpus.
\newblock In {\em Proceedings of the 21st ACM International Conference on
  Information and Knowledge Management}, CIKM '12, pages 1980--1984, New York,
  NY, USA, 2012. ACM.
\newblock 10.1145/2396761.2398556.

\bibitem{Yuan2016}
Shuhan Yuan, Xintao Wu, and Yang Xiang.
\newblock A two phase deep learning model for identifying discrimination from
  tweets.
\newblock In {\em Proceedings of 19th International Conference on Extending
  Database Technology}, pages 696--697, 2016.

\bibitem{Zhang2018}
Ziqi Zhang, David Robinson, and John Tepper.
\newblock Detecting hate speech on {T}witter using a convolution-{GRU} based
  deep neural network.
\newblock In {\em Proceedings of the 15th Extended Semantic Web Conference},
  ESWC'18, pages 745--760, Berline, Germany, 2018. Springer.
\newblock doi:10.1007/978-3-319-93417-4{\_}48.

\bibitem{Zhong2016}
Haoti Zhong, Hao Li, Anna Squicciarini, Sarah Rajtmajer, Christopher Griffin,
  David Miller, and Cornelia Caragea.
\newblock Content-driven detection of cyberbullying on the {Instagram} social
  network.
\newblock In {\em Proceedings of the Twenty-Fifth International Joint
  Conference on Artificial Intelligence}, IJCAI'16, pages 3952--3958, Menlo
  Park, California, United States, 2016. AAAI Press.

\end{thebibliography}


\begin{thebibliography}{5}
\ifx \bisbn   \undefined \def \bisbn  #1{ISBN #1}\fi
\ifx \binits  \undefined \def \binits#1{#1} \fi
\ifx \bauthor  \undefined \def \bauthor#1{#1} \fi
\ifx \bjtitle  \undefined \def \bjtitle#1{\textit{#1}}\fi
\ifx \batitle  \undefined \def \batitle#1{#1} \fi
\ifx \bctitle  \undefined \def \bctitle#1{#1} \fi
\ifx \bvolume  \undefined \def \bvolume#1{\textbf{#1}}\fi
\ifx \byear  \undefined \def \byear#1{#1} \fi
\ifx \bissue  \undefined \def \bissue#1{#1} \fi
\ifx \bfpage  \undefined \def \bfpage#1{#1} \fi
\ifx \blpage  \undefined \def \blpage #1{#1} \fi
\ifx \burl  \undefined \def \burl#1{#1} \fi
\ifx \doiurl  \undefined \def \doiurl#1{#1} \fi
\ifx \betal  \undefined \def \betal{et al.} \fi
\ifx \binstitute  \undefined \def \binstitute#1{#1} \fi
\ifx \beditor  \undefined \def \beditor#1{#1} \fi
\ifx \bpublisher  \undefined \def \bpublisher#1{#1} \fi
\ifx \bbtitle  \undefined \def \bbtitle#1{\textit{#1}} \fi
\ifx \bedition  \undefined \def \bedition#1{#1} \fi
\ifx \bseriesno  \undefined \def \bseriesno#1{#1} \fi
\ifx \blocation  \undefined \def \blocation#1{#1} \fi
\ifx \bsertitle  \undefined \def \bsertitle#1{#1} \fi
\ifx \bsnm \undefined \def \bsnm#1{#1} \fi
\ifx \bsuffix \undefined \def \bsuffix#1{#1} \fi
\ifx \bparticle \undefined \def \bparticle#1{#1} \fi
\ifx \barticle \undefined \def \barticle#1{#1} \fi
\ifx \botherref \undefined \def \botherref #1{#1} \fi
\ifx \url \undefined \def \url#1{#1} \fi
\ifx \bchapter \undefined \def \bchapter#1{#1} \fi
\ifx \bbook \undefined \def \bbook#1{#1} \fi
\ifx \bcomment \undefined \def \bcomment#1{#1} \fi
\ifx \oauthor \undefined \def \oauthor#1{#1} \fi
\ifx \citeauthoryear \undefined \def \citeauthoryear#1{#1} \fi
\ifx \texttildelow  \undefined \def \texttildelow{\symbol{126}} \fi
\def \endbibitem {}
\ifx \bconflocation  \undefined \def \bconflocation#1{#1} \fi

\bibitem{hankeknees}
\begin{barticle}
\bauthor{\binits{H.}~\bsnm{Hanke}} and
\bauthor{\binits{D.}~\bsnm{Knees}},
\batitle{A phase-field damage model based on evolving microstructure},
\bjtitle{Asymptotic Analysis}
\bvolume{101}
(\byear{2017}),
\bfpage{149}--\blpage{180}.
\end{barticle}
\endbibitem

\bibitem{lefever}
\begin{barticle}
\bauthor{\binits{E.}~\bsnm{Lefever}},
\batitle{A hybrid approach to domain-independent taxonomy learning},
\bjtitle{Applied Ontology}
\bvolume{11}(\bissue{3})
(\byear{2016}),
\bfpage{255}--\blpage{278}.
\end{barticle}
\endbibitem

\bibitem{meltzeretal}
\begin{bchapter}
\bauthor{\binits{P.S.}~\bsnm{Meltzer}},
\bauthor{\binits{A.}~\bsnm{Kallioniemi}} and
\bauthor{\binits{J.M.}~\bsnm{Trent}},
\bctitle{Chromosome alterations in human solid tumors},
in: \bbtitle{The Genetic Basis of Human Cancer},
\beditor{\binits{B.}~\bsnm{Vogelstein}} and
\beditor{\binits{K.W.}~\bsnm{Kinzler}}, eds,
\bpublisher{McGraw-Hill},
\blocation{New York},
\byear{2002},
pp.~\bfpage{93}--\blpage{113}.
\end{bchapter}
\endbibitem

\bibitem{murrayetal}
\begin{bbook}
\bauthor{\binits{P.R.}~\bsnm{Murray}},
\bauthor{\binits{K.S.}~\bsnm{Rosenthal}},
\bauthor{\binits{G.S.}~\bsnm{Kobayashi}} and
\bauthor{\binits{M.A.}~\bsnm{Pfaller}},
\bbtitle{Medical Microbiology},
\bedition{4th} edn,
\bpublisher{Mosby},
\blocation{St. Louis},
\byear{2002}.
\end{bbook}
\endbibitem

\bibitem{wilson}
\begin{botherref}
\oauthor{\binits{E.}~\bsnm{Wilson}},
Active vibration analysis of thin-walled beams,
PhD thesis,
University of Virginia,
1991.
\end{botherref}
\endbibitem

\end{thebibliography}


\begin{appendix}\label{embedding}
\section{Full results}



Our full results obtained with different word embeddings are shown in Table \ref{tab_full_results_baselines} for the three re-implemented state of the art methods, and Table \ref{tab_full_results_dnnplus} for the proposed CNN+sCNNN and CNN+GRU methods. \textbf{e.w2v}, \textbf{e.twt}, and \textbf{e.glv} each denotes the Word2Vec, Twitter, and GloVe embeddings respectively. We also analyse the percentage of OOV in each pre-trained word embeddings models, and show the statistics in Table \ref{tab_embeddings}. Note the figures are based on the datasets after applying the Twitter normaliser. Table \ref{tab_twitter_normaliser} shows the effect of normalisation, in terms of the coverage of hashtags in the embeddings on different datasets.

\begin{table}[thb]
    \centering
\caption{Percentage of hashtags covered in embedding models before and after applying the Twitter normalisation tool. B - before applying normalisation; A - after applying normalisation}
	\begin{tabular}{c|cc|cc|cc}
        \hline
        \multirow{2}{*}{\textbf{Dataset}} 		& \multicolumn{2}{c|}{\textbf{e.twt}} 	& 
           \multicolumn{2}{c|}{\textbf{e.w2v}}	& \multicolumn{2}{c}{\textbf{e.glv}} \\
           &B &A &B &A&B &A\\
				\hline
				WZ						& 37\%	& 91\%	& <1\% 	&44\%	&7\%	&89\%	\\ 
				WZ-S.amt				& 39\%	&89\%	& <1\% 	&73\%	&18\%	&87\%\\ 
				WZ-S.exp				& 39\%	&89\%	& <1\% 	&73\%	&18\%	&87\%\\ 
				WZ-S.gb					& 39\%	&89\%	& <1\% 	&73\%	&18\%	&87\%\\ 
				WZ.pj					& 38\%	&90\%	& <1\% 	&52\%	&10\%	&89\%\\ 
				DT						& 35\%	& 78\%	& <1\% 	&71\%	&25\%	&79\%\\ 
				RM						& 15\%	&90\%	& <1\% 	&83\%	&13\%	&89\%\\ 
				
				\hline
			    \end{tabular}
		
		\label{tab_twitter_normaliser}
\end{table}

\begin{table}[thb]
    \centering
\caption{Percentage of OOV in each pre-trained embedding model across all datasets.}
	\begin{tabular}{c|c|c|c}
        \hline
        \textbf{Dataset} 		& \textbf{e.twt} 	& \textbf{e.w2v}	& \textbf{e.glv} \\
				\hline
				WZ						& 4\%			& 13\% 	&6\%	\\ 
				WZ-S.amt				& 3\%			& 10\% 	&5\%	\\ 
				WZ-S.exp				& 3\%			& 10\% 	&5\%	\\ 
				WZ-S.amt				& 3\%			& 10\% 	&5\%	\\ 
				WZ.pj					& 4\%			& 14\% 	&7\%	\\ 
				DT						& 6\%			& 25\% 	&11\%	\\ 
				RM						& 4\%			& 11\% 	&6\%	\\ 
				
				\hline
			    \end{tabular}
		
		\label{tab_embeddings}
\end{table}

From the results, we cannot identify any word embeddings that consistently outperform others on all tasks and datasets, and there is also no strong correlation between the percentage of OOV in each word embeddings model and the obtained F1 with that model. Using the Gamback et al. baseline as an example (Table \ref{tab_full_results_baselines}), despite being the least complete embeddings model, e.w2v still obtained the best F1 when classifying racism tweets on 5 datasets. On the contrary, despite being the most complete embedding model, e.twt only obtained the best F1 when classifying sexism tweets on 3 datasets. 

Although counter-intuitive, this may not be very much a surprise, considering previous findings from \cite{Chiu16}. The authors showed that the performance of word embeddings on intrinsic evaluation tasks (e.g., word similarity) does not always correlate to that on extrinsic, or downstream tasks. In details, they showed that the context window size used to train word embeddings can affect the trade-off between capturing the domain relevance and functions of words. A large context window not only reduces sparsity by introducing more contexts for each word, it also better captures the topics of words. A small window on the other hand, captures word function. In sequence labelling tasks, they showed that word embeddings trained using a context window of just 1 performed the best. 

Arguably in our task, the topical relevance of words may be more important for the classification of hate speech. Although the Twitter based embeddings model has the best coverage, it may have suffered from insufficient context during training, since tweets are often very short compared to other corpora used to train Word2Vec and GloVe embeddings. As a result, the topical relevance of words captured by these embeddings may have lower quality than we expect, and therefore empirically, they do not lead to consistently better performance than Word2Vec or GloVe embeddings that are trained on very large, context rich, general purpose corpus.


\begin{table*}[thb]
    \centering
    	\caption{Full results obtained by all baseline models.}
    \begin{tabular}{p{1cm}|p{1.0cm}|p{0.2cm}p{0.2cm}p{0.2cm}|p{0.2cm}p{0.2cm}p{0.2cm}|p{0.2cm}p{0.2cm}p{0.2cm}|p{0.2cm}p{0.2cm}p{0.2cm}|p{.2cm}p{0.2cm}p{0.2cm}|p{.2cm}p{0.2cm}p{0.2cm}|p{0.2cm}p{0.2cm}p{0.2cm}}
        \hline
        	\multicolumn{2}{c|}{\multirow{2}{*}{\textbf{Dataset and classes}}}& \multicolumn{3}{c|}{\multirow{2}{*}{\textbf{SVM}}}	& 
        	\multicolumn{9}{c|}{\textbf{Gamback et al. \cite{Gamback2017}}}& 
        	\multicolumn{9}{c}{\textbf{Park et al. \cite{Park2017}}}\\\cline{6-23}
        	\multicolumn{2}{c|}{}&&&& \multicolumn{3}{c|}{e.w2v} & \multicolumn{3}{c|}{e.twt} & \multicolumn{3}{c|}{e.glv} & \multicolumn{3}{c|}{e.w2v} & \multicolumn{3}{c|}{e.twt} & \multicolumn{3}{c}{e.glv}\\\hline
        	
        	\multirow{4}{*}{WZ-S.amt}& 	& P &R &F & P &R &F & P &R &F & P &R &F & P &R &F & P &R &F & P &R &F\\\cline{2-23}
        			&racism 	&.17 &.44 &.22 &.56 &.14 &.22 &.41 &.06 &.10 &.44 &.03 &.03 &.49 &.10 &.17 &.36 &.04 &.08 &.56 &.03 &.05\\
        			&sexism 	&.59 &.79 &.68 &.84 &.76 &.80 &.86 &.73 &.79 &.86 &.76 &.81 &.84 &.77 &.80 &.86 &.75 &.80 &.87 &.76 &.81\\        			
        			&non-hate 	&.95 &.82 &.88 &.93 &.97 &.95 &.93 &.98 &.95 &.93 &.98 &.95 &.94 &.97 &.95 &.93 &.98 &.95 &.93 &.98 &.96\\\hline
    		
    		\multirow{3}{*}{WZ-S.exp}
    		&racism 	&.21 &.51 &.26  &.62 &.43 &.51 &.63  &.33 &.43 &.54 &.15 &.24 &.58 &.37 &.45 &.64 &.30 &.39 &.54 &.10 &.17\\
    		&sexism 	&.46 &.76 &.58  &.74 &.60 &.66 &.75  &.61 &.67 &.73 &.63 &.68 &.74 &.61 &.67 &.75 &.61 &.67 &.74 &.62 &.67\\    		
    		&non-hate 	&.96 &.83 &.89  &.94 &.97 &.95 &.94  &.97 &.95 &.94 &.97 &.95 &.94 &.97 &.95 &.94 &.97 &.95 &.94 &.97 &.95\\\hline
			
			\multirow{3}{*}{WZ-S.gb}
			&racism 	&.28 &.70 &.38 &.56 &.32 &.41 &.65 &.27 &.38 &.71 &.17 &.27 &.52 &.30 &.38 &.66 &.21 &.32 &.60 &.10 &.17\\
			&sexism 	&.54 &.79 &.64 &.81 &.72 &.76 &.84 &.71 &.77 &.81 &.72 &.76 &.81 &.72 &.76 &.81 &.72 &.76 &.81 &.73 &.76\\			
			&non-hate 	&.96 &.85 &.90 &.94 &.97 &.96 &.94 &.98 &.96 &.94 &.97 &.96 &.94 &.97 &.96 &.94 &.98 &.96 &.94 &.97 &.96\\\hline
			
			\multirow{3}{*}{WZ.pj}
			&racism 	&.54 &.68 &.60 &.75 &.66 &.70 &.74 &.67 &.70 &.76 &.65 &.70 &.75 &.64 &.69 &.74 &.65 &.70 &.75 &.68 &.72\\
			&sexism 	&.51 &.66 &.58 &.76 &.60 &.66 &.77 &.58 &.66 &.78 &.60 &.67 &.76 &.60 &.67 &.78 &.57 &.66 &.78 &.61 &.69\\			
			&non-hate 	&.86 &.75 &.80 &.84 &.91 &.88 &.84 &.92 &.88 &.84 &.92 &.88 &.84 &.91 &.88 &.84 &.92 &.88 &.85 &.91 &.88\\\hline
			
			\multirow{3}{*}{WZ}
			&racism 	&.54 &.65 &.59 &.74 &.67 &.70 &.72 &.73 &.72  &.75 &.72 &.73 &.74 &.70 &.72 &.74 &.72 &.73&.75 &.72 &.74\\
			&sexism 	&.51 &.66 &.57 &.76 &.61 &.66 &.73 &.62 &.67  &.77 &.59 &.67 &.76 &.61 &.66 &.74 &.60 &.66&.78 &.60 &.68\\			
			&non-hate 	&.85 &.75 &.79 &.84 &.90 &.87 &.85 &.89 &.87  &.85 &.91 &.88 &.85 &.91 &.88 &.85 &.90 &.87 &.85 &.91 &.88\\\hline
			
			\multirow{2}{*}{DT}
			&hate 	    &.15 &.52 &.23 &.45 &.20 &.28 &.49 &.21 &.30 &.54 &.25 &.34 &.48 &.20 &.28 &.51 &.22 &.31 &.55 &.24 &.33\\					
			&non-hate 	&.97 &.81 &.88 &.95 &.99 &.97 &.95 &.99 &.97 &.96 &.99 &.97 &.95 &.99 &.97 &.95 &.99 &.97 &.95 &.99 &.97\\\hline
			
			\multirow{2}{*}{RM}
			&hate 	    &.45 &.81  &.58 &.73 &.62 &.67 &.71 &.57 &.63 &.73 &.61 &.66 &.72 &.61 &.66 &.70 &.60 &.65 &.73 &.59 &.65\\					
			&non-hate 	&.96 &.78 &.86  &.92 &.95 &.94 &.92 &.95 &.93 &.92 &.95 &.94 &.92 &.95 &.94 &.92 &.95 &.93 &.92 &.95 &.94\\
				\hline

    \end{tabular}
	
		\label{tab_full_results_baselines}
\end{table*}


\begin{table*}[thb]
    \centering
    	\caption{Full results obtained by the CNN+GRU and CNN+sCNN models.}
    \begin{tabular}{p{1cm}|p{1.0cm}|p{0.31cm}p{0.31cm}p{0.31cm}|p{0.31cm}p{0.31cm}p{0.31cm}|p{0.31cm}p{0.31cm}p{0.31cm}|p{.25cm}p{0.31cm}p{0.31cm}|p{0.31cm}p{0.31cm}p{0.31cm}|p{0.31cm}p{0.31cm}p{0.31cm}}
        \hline
        	\multicolumn{2}{c|}{\multirow{2}{*}{\textbf{Dataset and classes}}}&  
        	\multicolumn{9}{c|}{\textbf{CNN+sCNN}}& 
        	\multicolumn{9}{c}{\textbf{CNN+GRU}}\\\cline{3-20}
        	\multicolumn{2}{c|}{}& \multicolumn{3}{c|}{e.w2v} & \multicolumn{3}{c|}{e.twt} & \multicolumn{3}{c|}{e.glv} & \multicolumn{3}{c|}{e.w2v} & \multicolumn{3}{c|}{e.twt} & \multicolumn{3}{c}{e.glv}\\\hline
        	
        	\multirow{4}{*}{WZ-S.amt}& 	& P &R &F & P &R &F & P &R &F & P &R &F & P &R &F & P &R &F\\\cline{2-20}
        			&racism 	&0.45 &0.22 &0.29 &0.43 &0.26 &0.32 &0.36 &0.21 &0.26 &0.47 &0.20 &0.28 &0.46 &0.14 &0.21 &0.45 &0.10 &0.16\\
        			&sexism 	&0.84 &0.78 &0.81 &0.80 &0.79 &0.79 &0.81 &0.81 &0.81 &0.80 &0.79 &0.80 &0.85 &0.73 &0.83 &0.79 &0.81 &0.80\\        			
        			&non-hate 	&0.94 &0.97 &0.95 &0.94 &0.96 &0.95 &0.95 &0.96 &0.95 &0.94 &0.96 &0.95 &0.93 &0.97 &0.95 &0.94 &0.97 &0.95\\\hline
    		
    		\multirow{3}{*}{WZ-S.exp}
    		&racism 	&0.60 &0.57 &0.58 &0.59  &0.67 &0.63 &0.52 &0.64 &0.57 &0.52 &0.48 &0.51 &0.57 &0.48 &0.52 &0.54 &0.38 &0.44\\
    		&sexism 	&0.72 &0.65 &0.68 &0.68  &0.76 &0.72 &0.67 &0.74 &0.70 &0.71 &0.65 &0.66 &0.71 &0.68 &0.69 &0.71 &0.65 &0.68\\    		
    		&non-hate 	&0.93 &0.95 &0.96 &0.96  &0.95 &0.95 &0.96 &0.95 &0.95 &0.94 &0.96 &0.95 &0.95 &0.96 &0.95 &0.94 &0.96 &0.95\\\hline
			
			\multirow{3}{*}{WZ-S.gb}
			&racism 	&0.55 &0.53 &0.54 &0.53 &0.63 &0.58 &0.50 &0.52 &0.51 &0.53 &0.46 &0.49 &0.60 &0.53 &0.56 &0.59 &0.40 &0.48\\
			&sexism 	&0.82 &0.73 &0.77 &0.78 &0.79 &0.79 &0.78 &0.80 &0.79 &0.76 &0.80 &0.78 &0.80 &0.75 &0.77 &0.78 &0.78 &0.78\\			
			&non-hate 	&0.94 &0.97 &0.96 &0.96 &0.96 &0.96 &0.96 &0.96 &0.96 &0.96 &0.96 &0.96 &0.95 &0.97 &0.96 &0.95 &0.96 &0.96\\\hline
			
			\multirow{3}{*}{WZ.pj}
			&racism 	&0.72 &0.73 &0.73 &0.64 &0.86 &0.74 &0.69 &0.80 &0.74 &0.74 &0.72 &0.73 &0.73 &0.70 &0.72 &0.71 &0.72 &0.71\\
			&sexism 	&0.75 &0.64 &0.69 &0.69 &0.71 &0.70 &0.70 &0.67 &0.68 &0.66 &0.69 &0.68 &0.73 &0.63 &0.68 &0.71 &0.64 &0.67\\			
			&non-hate   &0.86 &0.90 &0.88 &0.89 &0.84 &0.86 &0.87 &0.87 &0.87 &0.87 &0.86 &0.87 &0.85 &0.89 &0.87 &0.86 &0.88 &0.87\\\hline
			
			\multirow{3}{*}{WZ}
			&racism 	&0.73 &0.74 &0.74 &0.64 &0.87 &0.74  &0.69 &0.84 &0.76 &0.71 &0.74 &0.73 &0.74 &0.70 &0.72&0.71 &0.75 &0.73\\
			&sexism 	&0.79 &0.59 &0.68 &0.73 &0.63 &0.68  &0.73 &0.63 &0.68 &0.72 &0.63 &0.68 &0.74 &0.59 &0.66&0.75 &0.60 &0.66\\			
			&non-hate 	&0.85 &0.91 &0.88 &0.87 &0.86 &0.87  &0.87 &0.87 &0.87 &0.86 &0.88 &0.87 &0.84 &0.90 &0.87 &0.85 &0.89 &0.87\\\hline
			
			\multirow{2}{*}{DT}
			&hate 	    &0.51 &0.21 &0.30 &0.49 &0.35 &0.41 &0.50 &0.38 &0.43 &0.37 &0.25 &0.29 &0.44 &0.22 &0.30 &0.41 &0.31 &0.35\\					
			&non-hate 	&0.95 &0.99 &0.97 &0.96 &0.98 &0.97 &0.96 &0.98 &0.97 &0.96 &0.97 &0.97 &0.95 &0.98 &0.97 &0.96 &0.97 &0.97\\\hline
			
			\multirow{2}{*}{RM}
			&hate 	    &0.73 &0.69 &0.71 &0.64 &0.75 &0.69 &0.74 &0.65 &0.69 &0.71 &0.65 &0.68 &0.65 &0.66 &0.66 &0.66 &0.66 &0.66\\					
			&non-hate 	&0.94 &0.95 &0.94 &0.95 &0.91 &0.93 &0.93 &0.95 &0.94 &0.93 &0.95 &0.94 &0.94 &0.93 &0.94 &0.93 &0.93 &0.93\\
				\hline

    \end{tabular}
	
		\label{tab_full_results_dnnplus}
\end{table*}


\begin{figure*}
	\centering
	\includegraphics[width=400pt]{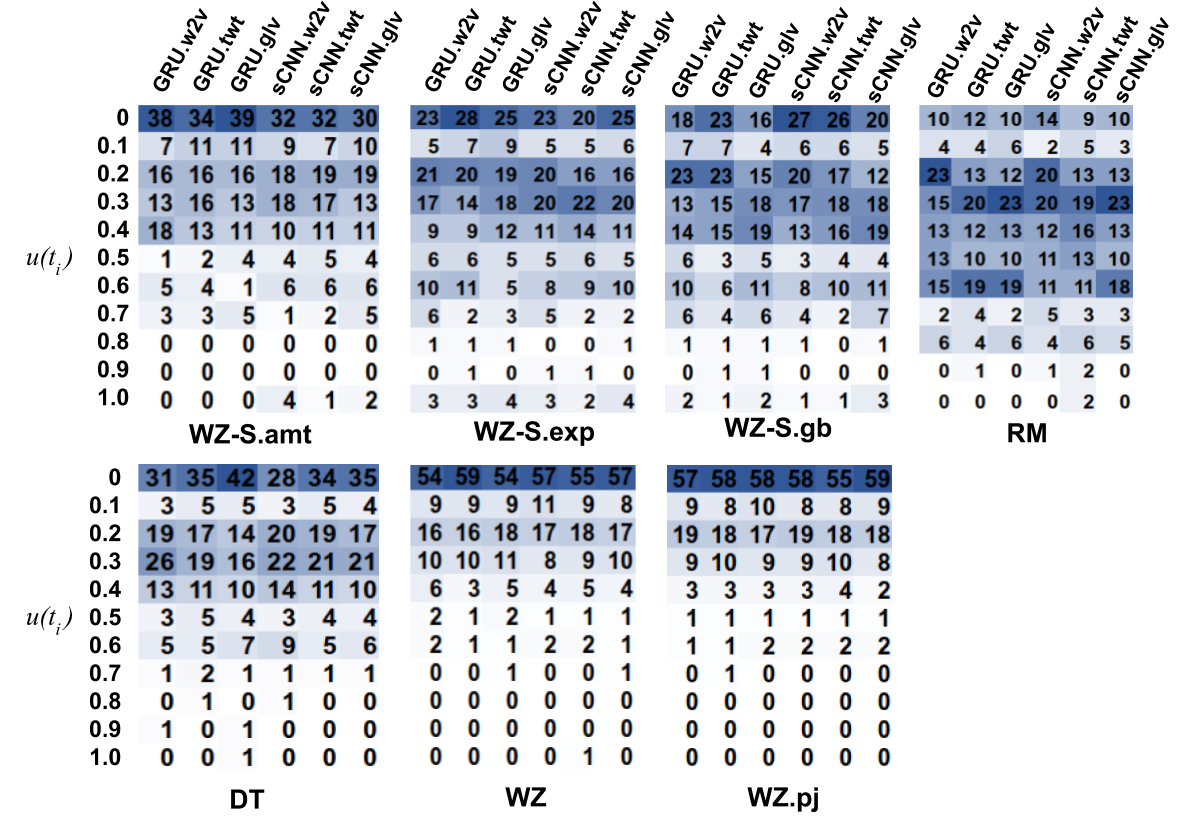}
	\caption{(Best viewed in colour) Distribution of additional true positives (compared against Gamback et al.) identified by CNN+sCNN (sCNN for shorthand) and CNN+GRU (GRU) over different ranges of uniqueness scores (see Equation \ref{e_uot}) on each dataset. Each row in a heatmap corresponds to a uniqueness score range. The numbers in each column sum up to 100\% while the colour scale within each cell is determined by the number label for that cell.}    
	\label{fig_heatmap_full} 
\end{figure*}

`\end{appendix}

\end{document}